\documentclass[runningheads]{llncs}
\usepackage{graphicx}
\usepackage{booktabs}
\usepackage{textcomp}
\usepackage{blindtext}
\usepackage{tikz}
\usepackage{comment}
\usepackage{amsmath} 
\usepackage{amssymb}
\usepackage{color}
\usepackage{cite} 
\usepackage{wrapfig}
\usepackage{setspace}
\usepackage{pifont}
\usepackage{multirow}
\usepackage{tabularx}
\usepackage{subcaption}
\usepackage[pagebackref,breaklinks,colorlinks]{hyperref}
\DeclareMathOperator*{\argmin}{argmin}
\DeclareMathOperator{\round}{round}
\usepackage[accsupp]{axessibility}  

\begin{document}
\pagestyle{headings}
\mainmatter
\def\ECCVSubNumber{}  

\title{TAFIM: Targeted Adversarial Attacks against Facial Image Manipulations} 

\titlerunning{TAFIM: Targeted Adversarial Attacks against Facial Image Manipulations}
\author{Shivangi Aneja\inst{1} \and
Lev Markhasin\inst{2} \and
Matthias Nie\ss ner \inst{1}}
\authorrunning{Aneja et al.}
\institute{Technical University of Munich, Germany
\and
Sony Europe RDC Stuttgart, Germany\\
}
\maketitle

\setcounter{footnote}{0}

\begin{figure}
    \centering
    \includegraphics[width=1.0\linewidth]{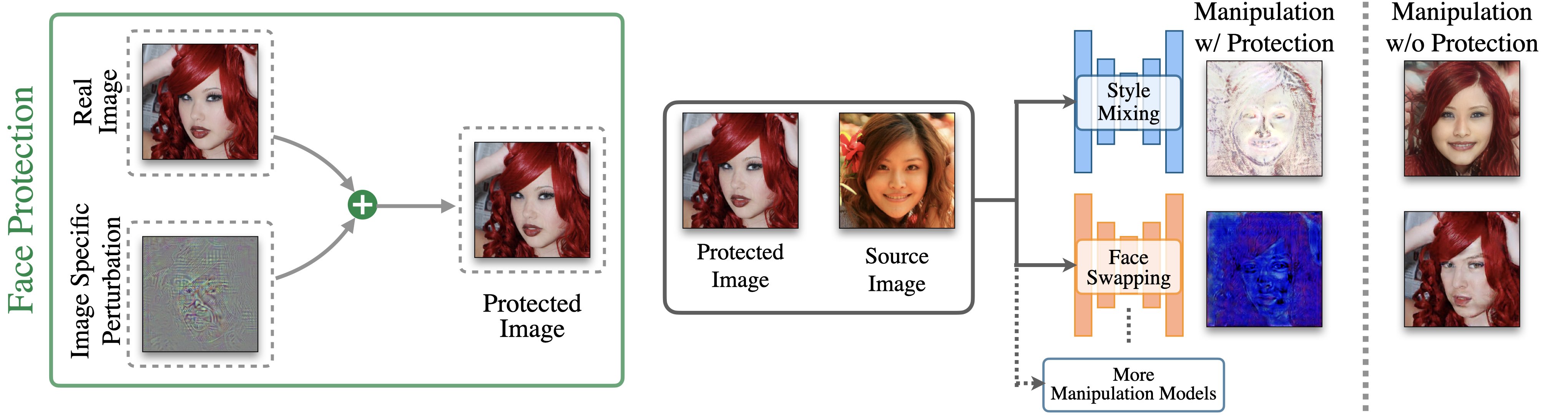}
    \vspace{-0.8cm}
    \caption{
     \textbf{Left}: We propose a novel approach to protect facial images from several image manipulation models simultaneously.
     We leverage neural network to encode the generation of quasi-imperceptible perturbations for different manipulation models and fuse them together using attention mechanism to generate manipulation-agnostic perturbation. This perturbation, when added to the real image, forces the face manipulation models to produce a predefined manipulation target as output (white/blue image in this case).
     This is several orders of magnitude faster and can also be used for real-time applications. \textbf{Right:} Without any protection applied, manipulation models can be misused to generate fake images for malicious activities. 
     } 
    \label{fig:teaser}
\end{figure}

%

\vspace{-1cm}
\begin{abstract}
Face manipulation methods can be misused to affect an individual's privacy or to spread disinformation. 
To this end, we introduce a novel data-driven approach that produces image-specific perturbations which are embedded in the original images.
The key idea is that these protected images prevent face manipulation by causing the manipulation model to produce a predefined manipulation target (uniformly colored output image in our case) instead of the actual manipulation.
In addition, we propose to leverage differentiable compression approximation, hence making generated perturbations robust to common image compression.
In order to prevent against multiple manipulation methods simultaneously, we further propose a novel attention-based fusion of manipulation-specific perturbations.
Compared to traditional adversarial attacks that optimize noise patterns for each image individually, our generalized model only needs a single forward pass, thus running orders of magnitude faster and allowing for easy integration in image processing stacks, even on resource-constrained devices like smartphones
\footnote{Project Page: \url{https://shivangi-aneja.github.io/projects/tafim}}.
\end{abstract}
\section{Introduction}
\label{sec:intro}

The spread of disinformation on social media has raised significant public attention in the recent few years, due to its implications on democratic processes and society in general.
The emergence and constant improvement of generative models, and in particular face image manipulation methods, has signaled a new possible escalation of this problem. 
For instance, face-swapping methods~\cite{ChenCNG20, nirkin2019fsgan} whose models are publicly accessible can be misused to generate non-consensual synthetic imagery. 
Other examples include face attribute manipulation methods ~\cite{richardson2021encoding, Patashnik2021ICCV, choi2018stargan, choi2020starganv2} that change the appearance of real photos, thus generating fake images that might then be used for criminal activities~\cite{fake_account}. 
Although a variety of manipulation tools have been open-sourced, surprisingly only a handful of methods have achieved widespread applicability among users (for details see the supplemental material). 
One reason is that re-training these methods is not only compute intensive but they also require specialized knowledge and skill sets for training.
As a result, most end users only apply easily accessible pre-trained models of a few popular methods. 
In this work, we exploit these popular manipulation methods and their models which are known in advance and propose targeted adversarial attacks to protect against facial image manipulations.

As powerful face image manipulation tools became easier to use and more widely available, many efforts to detect image manipulations were initiated by the research community~\cite{dolhansky2020deepfake}.
This has led to the task of automatically detecting manipulations as a classification task where predictions indicate whether a given image is real or fake. 
Several learning-based approaches~\cite{roessler2019faceforensicspp, Nguyen_2019, face_warping, deepfake_inconsistent_head_pose, Zhou_2017, mesonet, agarwal_protecting_2019, li2020face, cozzolino2018forensictransfer,  Cozzolino_2021_ICCV, aneja2020generalized} have shown promising results in identifying manipulated images. 
Despite the success and high classification accuracies of these methods, they can only be helpful if they are actually being used by the end-user.
However, manipulated images typically spread in private groups or on social media sites where manipulation detection is rarely available. 

An alternative avenue to detecting manipulations is to prevent manipulations from happening in the first place by disrupting potential manipulation methods~\cite{Yeh2020DisruptingID, ruiz2020disrupting, Yeh_2021_ICCV, yang2021faceguard, huang2021initiative}.
Here, the idea is to disrupt generative neural network models with low-level noise patterns, similar to the ideas of adversarial attacks used in the context of classification tasks~\cite{szegedy2014intriguing,goodfellow2015explaining}.
Methods optimizing noise patterns for every image from scratch~\cite{Yeh2020DisruptingID, ruiz2020disrupting, Yeh_2021_ICCV, yang2021faceguard} require several seconds to process a single image. 
In practice, this slow run time largely prohibits their use on mobile devices (e.g., as part of the camera stack).
Recently, Huang et al~\cite{huang2021initiative} introduced a neural network based approach to generate image-specific patterns for low-resolution images, however, they do not consider compression a common practical scenario and only consider a single manipulation model at a time, thus limiting its practical applicability.
At the same time, these manipulation prevention methods aim to either disrupt~\cite{9360910, ruiz2020disrupting, Li2019HidingFI,huang2021initiative} or nullify~\cite{Yeh2020DisruptingID,Yeh_2021_ICCV} the results of image manipulation models, which makes it difficult to identify which face manipulation technique was used.

To address these challenges, we propose a targeted adversarial attack against face image manipulation methods.
More specifically, we introduce a data-driven approach that generates quasi-imperceptible perturbations specific to a given image.
Our objective is that when an image manipulation is attempted, a predefined manipulation target is generated as output instead of the originally intended manipulation.
In contrast to previous optimization-based approaches, our perturbations are generated by a generalizable conditional model requiring only a few milliseconds for generation.
As a result, perturbations can be generated in only a few milliseconds rather than the required several seconds for existing works.
We additionally incorporate a differentiable compression module during training, to achieve robustness against common image processing pipelines. Finally, to handle multiple manipulation models simultaneously, we propose a novel attention-based fusion mechanism to combine model-specific perturbations.
In summary, the contributions in the paper are:

\begin{itemize}
    \item A data-driven approach to synthesize image-specific perturbations that outputs a predefined manipulation target (depending on the manipulation model used), instead of per-image optimization; this is not only significantly faster but also outperforms existing methods in terms of image-to-noise quality.
    \item Incorporation of differentiable compression during training to achieve robustness to common image processing pipelines.
    \item An attention-based fusion and refinement of model-specific perturbations to prevent against multiple manipulation models simultaneously.
\end{itemize}

\section{Related Work}
\textbf{Image Manipulation.}
Recent advances in image synthesis models have made it possible to generate detailed and expressive human faces ~\cite{thies2019neural,brock2018large, gaugan, karras2018progressive,karras2019stylebased, Karras_2020_CVPR,Karras2021} which might be used for unethical activities/frauds. Even more problematic can be the misuse of real face images to synthesize new ones. For instance, face-attribute modification techniques~\cite{Patashnik2021ICCV, richardson2021encoding, choi2018stargan, choi2020starganv2} and face-swapping models~\cite{nirkin2019fsgan,ChenCNG20} facilitate the manipulation of existing face images. 
Similarly, facial re-enactment tools~\cite{Zakharov19,Siarohin_2019_NeurIPS, thies2016face, kim2018deepvideo, Gafni_2021_CVPR} also use real images/videos to synthesize fake videos.

\textbf{Facial Manipulation Detection.} 
The increasing availability of these image manipulation models calls for the need to reliably detect synthetic images in an automated fashion. 
Traditional facial manipulation detection leverages handcrafted features such as gradients or compression artifacts, in order to find inconsistencies in an image~\cite{8267641,LyuPZ14,6210378}. 
While such self-consistency can produce good results, these methods are less accurate than more
recent learning-based techniques~\cite{roessler2019faceforensicspp, Cozzolino_2021_ICCV, Agarwal2019ProtectingWL, aneja2020generalized, cozzolino2018forensictransfer}, which are able to detect fake imagery with a high degree of confidence. 
In contrast to detecting forgeries, we aim to prevent manipulations from happening in the first place by rendering the respective manipulation models ineffective by introducing targeted adversarial attacks.

\textbf{Adversarial Attacks.} 
Adversarial attacks were initially introduced in the context of classification tasks~\cite{szegedy2014intriguing, goodfellow2015explaining,MoosaviDezfooli2016DeepFoolAS,Dong2018BoostingAA} and eventually expanded to semantic segmentation and detection models~\cite{fischer2017adversarial,xie2017adversarial, poursaeed2018generative}. 
The key idea behind these methods is to make imperceptible changes to an image in order to disrupt the feature extraction of the underlying neural networks. 
%
%
While these methods have achieved great success in fooling state-of-the-art vision models, one significant drawback is that optimizing a pattern for every image individually makes the optimization process quite slow. 
In order to address this challenge, generic universal image-agnostic noise patterns were introduced~\cite{Moosavi-Dezfooli_2017_CVPR, metzen2017universal}.
This has shown to be effective for misclassification tasks but gives suboptimal results for generative models, as we show in Sec.~\ref{sec:results}. 

\textbf{Manipulation Prevention.} 
Deep steganography and watermarking techniques~\cite{Zhu2018HiDDeNHD,2019stegastamp, wengrowski2019light, Luo2020DistortionAD, wang2021faketagger, yang2021faceguard} can be used to embed an image-specific watermark to secure an image. 
For instance, FaceGuard~\cite{yang2021faceguard} embeds a binary vector to the original image representative of a person's identity and classifies whether the image is fake by checking if the watermark is intact after being used for face manipulation tasks. 
These methods, however, cannot prevent the manipulation of face images which is the key focus of our work.

Recent works that aim to prevent image manipulations exploit adversarial attack techniques to break image manipulation models. 
Ruiz et al.~\cite{ruiz2020disrupting} disrupt the output of deepfake generation models. 
Yeh et al.~\cite{Yeh2020DisruptingID, Yeh_2021_ICCV} aim to nullify the effect of image manipulation models. 
Other approaches~\cite{Li2019HidingFI,9360910} aim to disturb the output of face detection and landmark extraction steps, which are usually used as pre-processing by deepfake generation methods. 
One commonality of these methods is that they optimize a pattern for each image separately which is computationally very expensive, thus having limited applicability for real-world applications like resource-constrained devices. 
Very recently, Huang et al~\cite{huang2021initiative} proposed a neural network based approach to generate image-specific patterns for low-resolution images, however, they do not consider compression, which is a common practical scenario that can make these generated patterns ineffective (as shown in Sec.~\ref{sec:results}). Additionally, this method only considers a single manipulation model at a time, thus limiting its applicability to protect against multiple manipulations simultaneously.
To this end, we propose (a) a novel data-driven method to generate image-specific perturbations which are robust to compression and (b) fusion of manipulation-specific perturbations. Our method not only require less computational effort compared to existing adversarial attacks works, but can protect against multiple manipulation methods simultaneously.

\begin{figure*}[t!]
\begin{center}
\includegraphics[width=\linewidth]{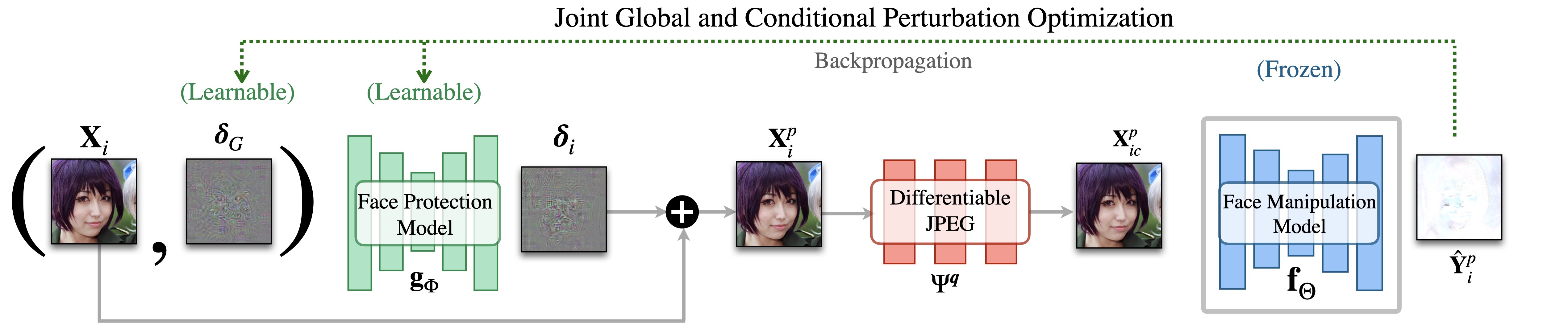}
\end{center}
\vspace{-0.5cm}
   \caption{ We first pass the real image $\textbf{X}_i$ and the global perturbation $\boldsymbol{\delta}_G$ through the face protection model $\textbf{g}_{\boldsymbol{\Phi}}$ to generate the image-specific perturbation $\boldsymbol{\delta}_i$. This perturbation is then added to the original image to create the protected image $\mathbf{X}_i^{p}$. The protected image is then compressed using the differentiable JPEG $\boldsymbol{\Psi^q}$ that generates compressed protected image $\mathbf{X}_{ic}^{p}$, which is passed through face manipulation model $\textbf{f}_{\boldsymbol{\Theta}}$ to generate the manipulated output $\hat{\textbf{Y}}_{i}^{p}$. The output of the face manipulation model is then used to drive the optimization.}
\label{fig:method_pipeline_single}
\end{figure*}

\begin{figure*}[t!]
\begin{center}
\includegraphics[width=\linewidth]{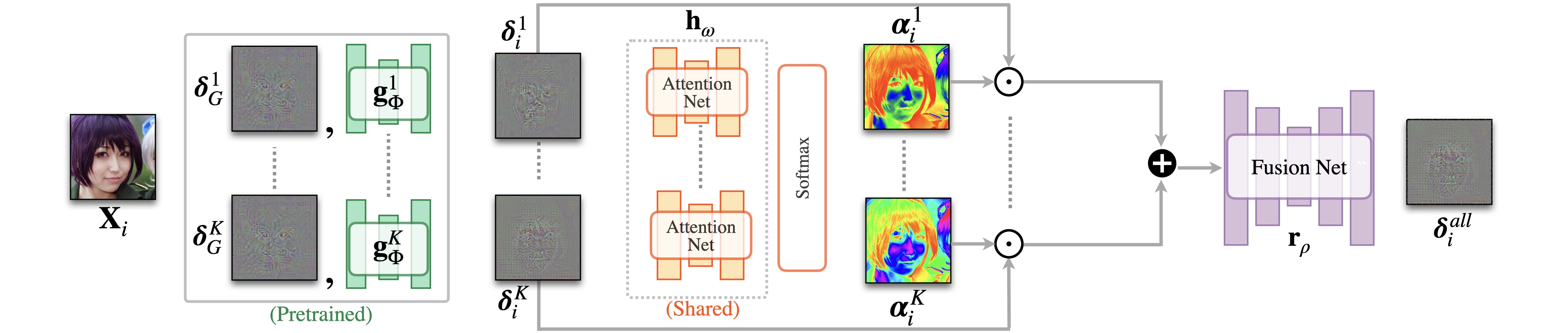}
\end{center}
\vspace{-0.5cm}
   \caption{For a given RGB image $\textbf{X}_i$, we first use the pre-trained manipulation-specific global noise and protection models $\{\boldsymbol{\delta}_G^k, \textbf{g}_{\boldsymbol{\Phi}}^{k}\}_{k=1}^{K}$ to generate manipulation-specific perturbations $\{\boldsymbol{\delta}_i^k\}_{k=1}^{K}$, which are passed into a shared attention backbone $\boldsymbol{h_\omega}$ to generate the spatial attention maps $\{\boldsymbol{\alpha}_i^k\}_{k=1}^{K}$. These attention maps are then combined with manipulation-specific $\{\boldsymbol{\delta}_i^k\}_{k=1}^{K}$ using channel-wise hadamard product and blended together using addition operation. Finally, the blended perturbation is then refined using FusionNet $\boldsymbol{r_\rho}$ to generate manipulation-agnostic perturbation $\boldsymbol{\delta}_i^{\text{all}}$.}
\label{fig:method_pipeline_multiple}
\end{figure*}

\section{Proposed Method}
Our goal is to prevent face image manipulations and simultaneously identify which model was used for the manipulation. 
That is, for a given face image, we aim to find an imperceptible perturbation that disrupts the generative neural network of a manipulation method such that a solid color image is produced as output instead of originally-intended manipulation.
Algorithmically, this is a targeted adversarial attack where the predefined manipulation targets make it easy for a human to identify the used manipulation method.

\subsection{Method Overview}
We consider a setting where we are given $K$ manipulation models $\mathcal{M} = \{\textbf{f}^k_{\boldsymbol{\Theta}}\}_{k=1}^{K}$ where $\textbf{f}^k_{\boldsymbol{\Theta}}$ denotes the $k$-th manipulation model.
For a given RGB image $\mathbf{X}_i \in \mathbb{R}^{H \times W \times 3}$ of height $H$ and width $W$, the goal is to find the optimal perturbation $\boldsymbol{\delta}_i \in \mathbb{R}^{H \times W \times 3}$ that is embedded in the original image $\mathbf{X}_i$ to produce a valid protected image $\mathbf{X}_i^p\in \mathbb{R}^{H \times W \times 3}$.
The manipulation model $\textbf{f}^k_{\boldsymbol{\Theta}}$, which is parametrized by its neural network weights $\boldsymbol{\Theta}$, is also given as input to the method. 
Note that we use $\textbf{f}^k_{\boldsymbol{\Theta}}$ only to drive the perturbation optimization and do not alter its weights. 
For the given image $\textbf{X}_i$, the output synthesized by the manipulation model $\textbf{f}^k_{\boldsymbol{\Theta}}$ is denoted as $\mathbf{\hat{Y}}_{ik} \in \mathbb{R}^{H \times W \times 3}$. 
We define the uniformly-colored predefined manipulation targets for the $K$ manipulation models as $\mathcal{Y} = \{ \mathbf{Y}^{\text{target}}_k \}_{k=1}^{K}$.

In order to protect face images and obtain image perturbations, we propose two main ideas: 
First, for a given manipulation model $\textbf{f}_{\boldsymbol{\Theta}}$, we jointly optimize for a global perturbation pattern $\boldsymbol{\delta}_G \in \mathbb{R}^{H \times W \times 3}$ and a generative neural network $\textbf{g}_{\boldsymbol{\Phi}}$ (parameterized by its weights $\boldsymbol{\Phi}$) to produce image-specific perturbations $\boldsymbol{\delta}_i$. 
The global pattern $\boldsymbol{\delta}_G$ is generalized across the entire data distribution. 
The generative model $\boldsymbol{g_{\Phi}}$ is conditioned on the global perturbation $\boldsymbol{\delta}_G$ as well as the real image $\mathbf{X}_i$. 
Our intuition is that the global perturbation provides a strong prior for the global noise structure, thus enabling the conditional model to produce more effective perturbations. 
We also incorporate a differentiable JPEG module to ensure the robustness of the perturbations towards compression. This is shown in Fig.~\ref{fig:method_pipeline_single}.

Second, to handle multiple manipulation models simultaneously, we leverage an attention network $\boldsymbol{h_{\omega}}$ (parametrized by $\boldsymbol{\omega}$) to first generate attention maps $\{\alpha_{k}\}_{k=1}^{K}$ for the $K$ manipulation methods, which are then used to refine the model-specific perturbations $ \{\boldsymbol{\delta}_{i}^k\}_{k=1}^{K}$ with an encoder-decoder network denoted as FusionNet $\boldsymbol{r_{\rho}}$ (parametrized by $\boldsymbol{\rho}$) to generate a single final perturbation $\boldsymbol{\delta}_{i}^{\text{all}}$ for the given image $\mathbf{X}_i$. $\boldsymbol{\delta}_{i}^{\text{all}}$ can protect the image from the given K manipulation methods simultaneously. 
An overview is shown in Fig.~\ref{fig:method_pipeline_multiple}.

\subsection{Methodology}
We define an optimization strategy where the objective is to find the smallest possible perturbation that achieves the largest disruption in the output manipulation; i.e., where the generated output for the $k$-th manipulation model is closest to its predefined target image $\textbf{Y}^{\text{target}}_k$. This is explained in detail below.

\subsubsection{Joint Global and Conditional Generative Model Optimization  }
The global perturbation $\boldsymbol{\delta}_G \in \mathbb{R}^{H \times W \times 3} $ is a fixed image-agnostic perturbation shared across the data distribution.
The conditional generative neural network model $\boldsymbol{g_{\Phi}}$ is a UNet~\cite{Ronneberger2015UNetCN} based encoder-decoder architecture. 
For a given manipulation method, we jointly optimize global perturbation $\boldsymbol{\delta}_G$  and the parameters $\boldsymbol{\Phi}$ of this conditional model $\boldsymbol{g_{\Phi}}$ together in order to generate image-specific perturbations. 
\begin{equation}
    \boldsymbol{\delta}^*_G, \boldsymbol{\Phi^*} =\argmin_{\delta_G, \Phi} \ \ \mathcal{L}_k
\end{equation}
where $\mathcal{L}_k$ refers to overall loss (Eq.~\ref{eq:total_loss}).
\begin{equation}
    \mathcal{L}_k = \Biggl[ \sum\limits_{i=1}^{N}  \mathcal{L}^{\text{recon}}_i + \lambda \mathcal{L}^{\text{perturb}}_i \Biggl]_k ,
\label{eq:total_loss}
\end{equation}
where the parameter $\lambda$ regularizes the strength of perturbation added to the real image, $N$ denotes the number of images in the dataset, $i$ denotes the image index and $k$ denotes the manipulation method. $\mathcal{L}^{\text{recon}}_i$ and $\mathcal{L}^{\text{perturb}}_i$ represent reconstruction and perturbation losses for $i$-th image.
The model $\boldsymbol{g_{\Phi}}$ is conditioned on the globally-optimized perturbation $\boldsymbol{\delta_G}$ as well as the original input image $\mathbf{X}_i$. 
Conditioning the model $\boldsymbol{g_{\Phi}}$ on $\boldsymbol{\delta}_G$ facilitates the transfer of global structure from the facial imagery to produce highly-efficient perturbations, i.e., these perturbations are more successful in disturbing manipulation models to produce results close to the manipulation targets. 
The real image $\mathbf{X}_i$ and global perturbation $\boldsymbol{\delta}_G$ are first concatenated channel-wise, $\widehat{\mathbf{X}}_i = \big[\mathbf{X}_i, \boldsymbol{\delta}_G\big]$, to generate a six-channel input $ \widehat{\mathbf{X}}_i \in \mathbb{R}^{H \times W \times 6}$.

$\widehat{\mathbf{X}}_i$ is then passed through the conditional model $\boldsymbol{g_{\Phi}}$ to generate image-specific perturbation $\boldsymbol{\delta}_i = \boldsymbol{g_{\Phi}} (\widehat{\mathbf{X}}_i)$.
These image-specific perturbations $\boldsymbol{\delta}_i$ are then added to the respective input images $\mathbf{X}_i$ to generate the protected image $\mathbf{X}_i^{p}$ as
\begin{equation}
\label{eq:protected_img_specific}
 \mathbf{X}_i^{p} = \text{Clamp}_{\varepsilon} (\mathbf{X}_i + \boldsymbol{\delta}_i) .
\end{equation}
The  $\text{Clamp}_{\varepsilon} (\xi)$ function projects higher/lower values of $\xi$ into the valid interval $[-\varepsilon, \varepsilon]$. 
Similarly, we generate the protected image using global perturbation $\boldsymbol{\delta}_G$ as
\begin{equation}
\label{eq:protected_img_global}
 \mathbf{X}_i^{Gp} = \text{Clamp}_{\varepsilon} (\mathbf{X}_i + \boldsymbol{\delta}_G) .
\end{equation}
For the generated conditional and global protected image $\mathbf{X}_i^{p}$ and $\mathbf{X}_i^{Gp}$ and the given manipulation model $\textbf{f}^k_{\boldsymbol{\Theta}}$, the reconstruction loss $\mathcal{L}^{\text{recon}}_i$ and perturbation loss $\mathcal{L}^{\text{perturb}}_i$ are formulated as 

\begin{equation}
    \mathcal{L}^{\text{recon}}_i = \Big\| \textbf{f}^k_{\boldsymbol{\Theta}}(\mathbf{X}_i^{p}) - \mathbf{Y}_k^{\text{target}} \Big\|_2 + \Big\| \textbf{f}^k_{\boldsymbol{\Theta}}(\mathbf{X}_i^{Gp}) - \mathbf{Y}_k^{\text{target}} \Big\|_2 .
\end{equation}

\begin{equation}
    \mathcal{L}^{\text{perturb}}_i = \Big\| \mathbf{X}_i^{p} -\mathbf{X}_i  \Big\|_2 + \Big\| \mathbf{X}_i^{Gp} -\mathbf{X}_i  \Big\|_2 .
\label{eq:perturb_loss}
\end{equation}

Finally, the overall loss can then be written as
\begin{equation}
\begin{split}
    \mathcal{L}_k =  \Biggl[ \sum\limits_{i=1}^{N} \Big\| \textbf{f}^k_{\boldsymbol{\Theta}}(\mathbf{X}_i^{p}) - \mathbf{Y}_k^{\text{target}} \Big\|_2 + \Big\| \textbf{f}^k_{\boldsymbol{\Theta}}(\mathbf{X}_i^{Gp}) - \mathbf{Y}_k^{\text{target}} \Big\|_2 + \\
    \lambda \Bigg( \Big\| \mathbf{X}_i^{p} -\mathbf{X}_i \Big\|_2 + \Big\| \mathbf{X}_i^{Gp} -\mathbf{X}_i  \Big\|_2 \Bigg)  \Biggl]_k .
\end{split}
\end{equation}
The global perturbation $\boldsymbol{\delta}_G$ is initialized with a random vector sampled from a multivariate uniform distribution, i.e., $\boldsymbol{\delta}^0_\mathbf{G} \sim \mathcal{U}(\textbf{0},\textbf{1})$ and optimized iteratively. Note that $\mathbf{X}_i^{Gp}$ is used only to drive the optimization of $\boldsymbol{\delta}_G$. For further details on the network architecture and hyperparameters, we refer to Sec.~\ref{sec:results} and the supplemental material.

\subsubsection{Differentiable JPEG Compression}
In many practical scenarios, images shared on social media platforms get compressed over the course of transmission. 
Our initial experiments suggest that protected images $\mathbf{X}_i^{p}$ generated from the previous steps can easily become ineffective by applying image compression. 
In order to make our perturbations robust, we propose to incorporate a differentiable JPEG compression into our generative model; i.e., we aim to generate perturbations that still disrupt the manipulation models even if the input is compressed.
The actual JPEG compression technique~\cite{wallace1992jpeg} is non-differentiable due to the lossy quantization step (details in supplemental) where information loss happens with the $\round$ operation as, $x:= \round(x).$ 
Therefore, we cannot train our protected images against the original JPEG technique. 
Instead, we leverage continuous and differentiable approximations~\cite{Shin2017JPEGresistantAI,Korus_2019_CVPR} to the rounding operator. 
For our experiments, we use the $\sin$ approximation by Korus et al.~\cite{Korus_2019_CVPR}
\begin{equation}
    x := x - \dfrac{\sin(2\pi x)}{2\pi} .
\end{equation}

This differentiable $\round$ approximation coupled with other transformations from the actual JPEG technique can be formalized into differentiable JPEG operation. 
We denote the full differentiable JPEG compression as $\boldsymbol{\Psi^q}$, where $\boldsymbol{q}$ denotes the compression quality. 

For training, we first map the protected image $\mathbf{X}_i^{p}$ to RGB colorspace $[0, 255]$ before applying image compression, obtaining $\widetilde{\mathbf{X}}_i^{p}$. 
Next, the image $\widetilde{\mathbf{X}}_i^{p}$ is passed through differential JPEG layers $\boldsymbol{\Psi^q}$ to generate a compressed image $\widetilde{\mathbf{X}}_{ic}^{p}$, which is then normalized again as $\mathbf{X}_{ic}^{p}$ before passing it to the manipulation model $\textbf{f}_\mathbf{{\Theta}}$.

Training with a fixed compression quality ensures robustness to that specific quality but shows limited performance when evaluated with different compression qualities. 
We therefore, generalize across compression levels by training our model with different compression qualities. 
Specifically, at each iteration, we randomly sample quality $\boldsymbol{q}$ from a discrete uniform distribution $\mathcal{U_D}(1,99)$, i.e. $\boldsymbol{q} \sim \mathcal{U_D}(1,99)$ and compress the protected image $\mathbf{X}_{i}^{p}$ at quality level $\boldsymbol{q}$.

This modifies the reconstruction loss $\mathcal{L}_{\text{recon}}$ as follows
\begin{equation}
    \mathcal{L}^{\text{recon}}_i = \Big\| \textbf{f}^k_{\boldsymbol{\Theta}}(\boldsymbol{\Psi^q} (\mathbf{X}^p_i)) - \mathbf{Y}^{\text{target}}_k \Big\|_2 + \Big\| \textbf{f}^k_{\boldsymbol{\Theta}}(\mathbf{X}_i^{Gp}) - \mathbf{Y}_k^{\text{target}} \Big\|_2
\end{equation}
where $\mathbf{X}_{ic}^{p} = \boldsymbol{\Psi^q} (\mathbf{X}^p_i)$ denotes the compressed protected image. Backpropagating the gradients through $\boldsymbol{\Psi^q}$ during training ensures that the added perturbations survive different compression qualities. 
At test time, we evaluate results with actual JPEG compression technique instead of approximated/differential used during training to report the results.

\subsubsection{Multiple Manipulation Methods}
To handle multiple manipulation models simultaneously, we combine model-specific perturbations $\{\boldsymbol{\delta}_i^k\}_{k=1}^{K}$ obtained previously using $\{\boldsymbol{\delta}_G^k, \textbf{g}_{\boldsymbol{\Phi}}^{k}\}_{k=1}^{K}$ and feed them to our attention network $ \textbf{h}_{\boldsymbol{\omega}}$ (parameterized by $\boldsymbol{\omega}$) and fusion network $ \textbf{r}_{\boldsymbol{\rho}}$ (parameterized by $\boldsymbol{\rho}$ ) to generate model-agnostic perturbations $\boldsymbol{\delta}_i^{\text{all}}$. 

\begin{equation}
    \boldsymbol{\omega^*}, \boldsymbol{\rho^*} =\argmin_{\omega, \rho} \ \ \mathcal{L}_{\text{all}} .
\end{equation} 

We leverage the pre-trained global pattern and conditional perturbation model pairs $\{\boldsymbol{\delta}_G^k, \textbf{g}_{\boldsymbol{\Phi}}^{k}\}_{k=1}^{K}$ for each of the $K$ different models to generate the final perturbation $\boldsymbol{\delta}_i^{\text{all}}$ for image $\mathbf{X}_i$. More precisely, for the image $\mathbf{X}_i$, we first use the pre-trained $\{\boldsymbol{\delta}_G^k, \textbf{g}_{\boldsymbol{\Phi}}^{k}\}_{k=1}^{K}$ to generate the model-specific perturbations $\{\boldsymbol{\delta}_i^k\}_{k=1}^{K}$ as:
\begin{equation}
\boldsymbol{\delta}_i^k = \textbf{g}_{\boldsymbol{\Phi}}^{k} (\mathbf{X}_i, \boldsymbol{\delta}_G^k).
\end{equation}
Next, these model-specific perturbations $\{\boldsymbol{\delta}_i^k\}_{k=1}^{K}$ are fed into attention module $\textbf{h}_{\boldsymbol{\omega}}$ coupled with the softmax operation to generate spatial attention maps $\{\boldsymbol{\alpha}_i^k\}_{k=1}^{K}$ as:
\begin{equation}
\boldsymbol{\alpha}_i^k = \frac{\text{exp} \big(\textbf{h}_{\boldsymbol{\omega}} (\boldsymbol{\delta}_i^k, C_k)\big)}{\sum\limits_{k=1}^{K} \text{exp} \big(\textbf{h}_{\boldsymbol{\omega}} (\boldsymbol{\delta}_i^k, C_k)\big)}.
\end{equation}
where $\boldsymbol{\alpha}_i^k \in \mathbb{R}^{H \times W} $ and $C_k$ refer to class label for the $k$-th manipulation model. These spatial attention maps are then blended with model-specific perturbations and refined with a fusion network $\textbf{r}_{\boldsymbol{\rho}}$ to generate the final perturbation $\boldsymbol{\delta}_i^{\text{all}}$ as:
\begin{equation}
\boldsymbol{\delta}_i^{\text{all}} = \textbf{r}_{\boldsymbol{\rho}}\Bigg( \sum\limits_{k=1}^{K} (\boldsymbol{\alpha}_i^k \odot \boldsymbol{\delta}_i^k)\Bigg)
\end{equation}
Finally, $\boldsymbol{\delta}_i^{\text{all}}$ is added to the image $\mathbf{X}_i$ to generate the common protected image  $\mathbf{X}_i^\text{all} = \text{Clamp}_{\varepsilon} (\mathbf{X}_i + \boldsymbol{\delta}_i^\text{all})$ and total loss is formalized as 
\begin{equation}
\begin{split}
    \mathcal{L}_{\text{all}} =  \sum\limits_{i=1}^{N} \Biggl[ \sum\limits_{k=1}^{K} \Bigg(    \Big\| \textbf{f}^k_{\boldsymbol{\Theta}}(\mathbf{X}_{i}^{\text{all}}) - \mathbf{Y}_k^{\text{target}} \Big\|_2 \Bigg) +
    \lambda \Big\| \boldsymbol{\delta}_i^{\text{all}}  \Big\|_2  \Biggl].
\end{split}
\end{equation}

\section{Results}~\label{sec:results}
We compare our method against well-studied adversarial attack baselines I-FGSM~\cite{Kurakin2017AdversarialML} and I-PGD~\cite{Madry2018TowardsDL}. 
To demonstrate our results, we perform experiments with three different models: (1) pSp Encoder~\cite{richardson2021encoding} which can be used for self-reconstruction and style-mixing (protected with solid white image as manipulation target), and (2) SimSwap~\cite{ChenCNG20} for face-swapping (protected with solid blue as manipulation target). (3) StyleClip~\cite{Patashnik2021ICCV} for text-driven manipulation (protected with solid red as manipulation target). 
For all these manipulations, we use the publicly available pre-trained models. 
For pSp encoder, we use a model that is trained for a self-reconstruction task. 
The same model can also be used for style-mixing to synthesize new images by mixing the latent style features of two images. 
For style-mixing and face-swapping, protection is applied to the target image. 
We introduce a custom split on FFHQ~\cite{karras2019stylebased} for our experiments. We use 10K images for training and 1K images for val and test split each. More details can be found in supplemental.
All results are reported on the corresponding test sets for each task respectively.

\begin{figure}[h!]
\begin{center}
\includegraphics[width=1.0\linewidth]{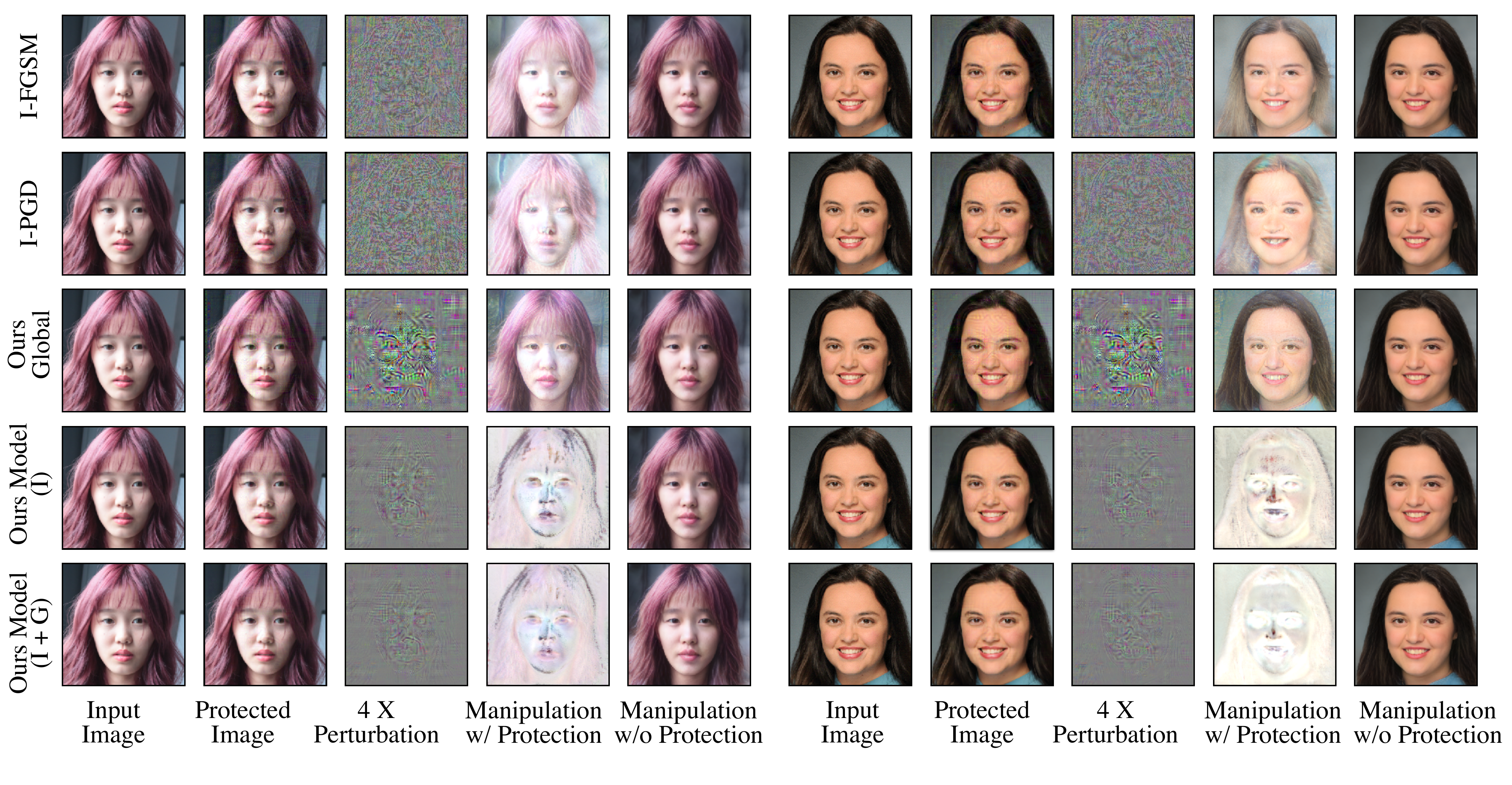}
\end{center}
\vspace{-0.8cm}
  \caption{Comparison on self-reconstruction task with white image as manipulation target. Perturbation enlarged ($4 \times$) for better visibility. \emph{Ours Global} refers to the optimized single global perturbation for all the images. \emph{Ours Model (I)} refers to the model conditioned only on real images and \emph{Ours Model (I + G)} refers to the model conditioned on global perturbation and real image, outperforms alternate baselines.}
\label{fig:uncompressed_results_self_recon}
\end{figure}

\textbf{Experimental Setup.} 
All images are first resized to $256 \times 256$ pixels. 
The global perturbation and conditional model are jointly optimized for 100k iterations with a learning rate of 0.0001 and Adam optimizer. 
For the protection $\boldsymbol{g_{\Phi}}$, attention $\boldsymbol{h_{\omega}}$ and fusion $\boldsymbol{r_{\rho}}$ network, we use the same UNet-64 encoder-decoder architecture.
We use a batch size of 1 for all our experiments. 
For I-PGD, we use a step size of 0.01. 
Both I-FGSM and I-PGD are optimized for 100 steps for every image in the test split. More details on training setup and hyperparameters can be found in the supplemental.

\textbf{Metrics.} 
To evaluate the output quality, we compute relative performance at different perturbation levels, i.e., we plot a graph with the x-axis showing different perturbation levels for the image and the y-axis showing how close is the output of the face manipulation model to the predefined manipulation target. 
We plot the graph for RMSE, PSNR, LPIPS~\cite{zhang2018perceptual} and VGG loss. 
In the optimal setting, for a low perturbation in the image, the output should look identical to the manipulation target; i.e., a lower graph is better for RMSE, LPIPS and VGG loss and higher for PSNR. 

\begin{figure}[h!]
\begin{center}
\includegraphics[width=1.0\linewidth]{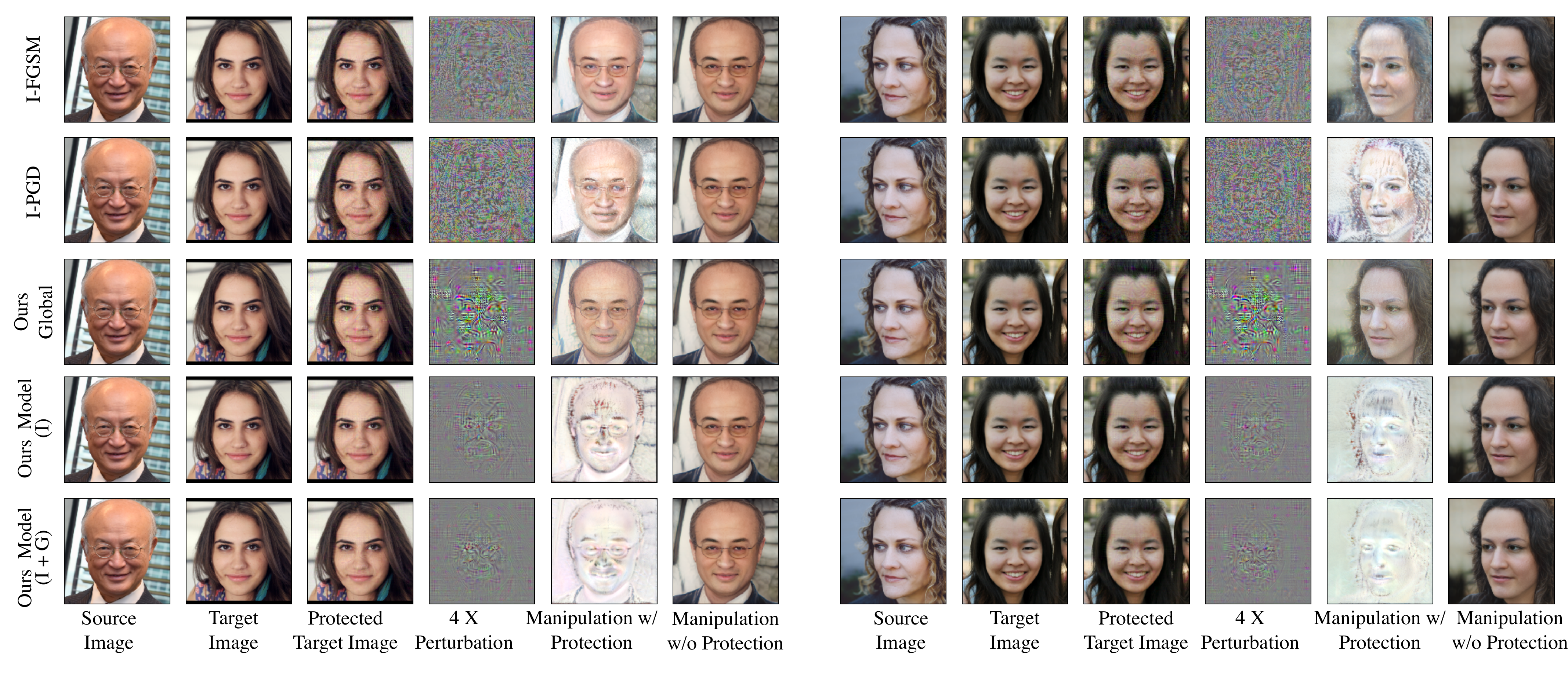}
\end{center}
\vspace{-0.8cm}
  \caption{Comparison on the style-mixing task (white target). The protection is applied to the target image. All methods are trained only for the self-reconstruction task and evaluated on style-mixing. Perturbation enlarged ($4 \times$) for better visibility. \emph{Ours Global} refers to the optimized single global perturbation. \emph{Ours Model (I)} refers to the model conditioned only on real images and \emph{Ours Model (I + G)} refers to the model conditioned on global perturbation and real image, outperforms alternate baselines.}
\label{fig:uncompressed_result_style_mix}
\end{figure}

\begin{figure}[h!]
\begin{center}
\includegraphics[width=1.0\linewidth]{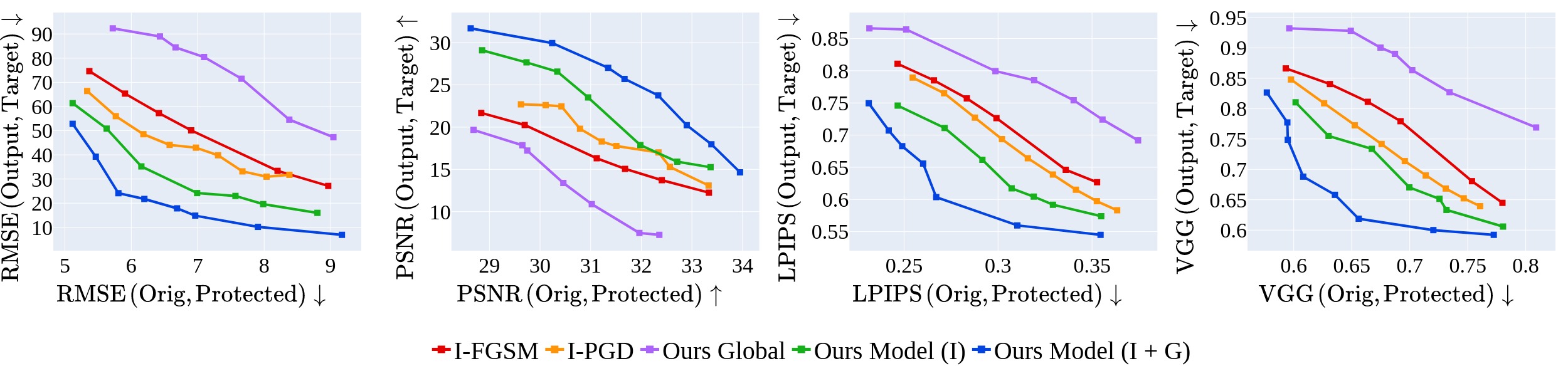}
\end{center}
\vspace{-0.7cm}
   \caption{Comparison with different optimization techniques evaluated on self-reconstruction (white target). We plot the output image quality (y-axis) corresponding to different levels of perturbations added to the image (x-axis). \emph{Orig} and \emph{Protected} refer to the original and protected image. \emph{Output} refers to the output of the manipulation model and \emph{Target} indicates the predefined manipulation target. Note that our method outperforms other baselines at all the different perturbation levels.}
\label{fig:perf_graph_self_recon}
\end{figure}

\textbf{Baseline Comparisons.} 
To compare our method against other adversarial attack baselines, we first evaluate the results of our proposed method on a single manipulation model without compression; i.e., neither training nor evaluating for JPEG compression. 
Visual results for self-reconstruction and style mixing are shown in Figs.~\ref{fig:uncompressed_results_self_recon} and~\ref{fig:uncompressed_result_style_mix}.
The performance graph for different perturbation levels is shown in Fig.~\ref{fig:perf_graph_self_recon}. 
We observe that the model conditioned on the global perturbation as well as real images outperforms the model trained only with real images, indicating that the global perturbation provides a strong prior in generating more powerful perturbations.

\textbf{Runtime Comparison.} 
We compare run-time performance against state-of-the-art in Tab.~\ref{tab:timing_comp}.
I-FGSM~\cite{Kurakin2017AdversarialML} and I-PGD~\cite{Madry2018TowardsDL} optimize for perturbation patterns for each image individually at run time; hence they are orders of magnitude slower than our method that only requires a single forward pass of our conditional generative neural network. Our model takes only $77.89 \pm 2.71$ ms and 117.0 MB memory to compute the perturbation for a single image on an Intel(R) Xeon(R) W-2133 CPU @ 3.60GHz. 
This is an order of magnitude faster compared to per-image methods that are run on GPUs. 
We believe this makes our method ideally suited to real-time scenarios, even on mobile hardware. 

\begin{table}[htpb]
      \centering
      \small
      \caption{Run-time performance (averaged over 10 runs) to generate  a perturbation for a single image on the self-reconstruction task. Our method runs an order of magnitude faster than existing works that require per-image optimization. All timings are measured on an Nvidia Titan RTX 2080 GPU.}
  \label{tab:timing_comp}
  \begin{tabular}{l| l  l}
  \toprule
 {\textbf{Method}} &  \textbf{Time} \\ 
\midrule
     {I-FGSM}~\cite{Kurakin2017AdversarialML}& 17517.71 ms & \footnotesize{\textcolor{gray}{($\pm124.08$ ms)}}  \\
     {I-PGD~\cite{Madry2018TowardsDL}}& 17523.01 ms & \footnotesize{\textcolor{gray}{($\pm204.15$ ms)}}  \\
     {Ours} & \textbf{10.66 ms} & \footnotesize{\textcolor{gray}{($\pm0.21$ ms)}} \\
    \bottomrule
  \end{tabular}
\end{table}

\begin{figure}[h!]
\begin{center}
\includegraphics[width=1.0\linewidth]{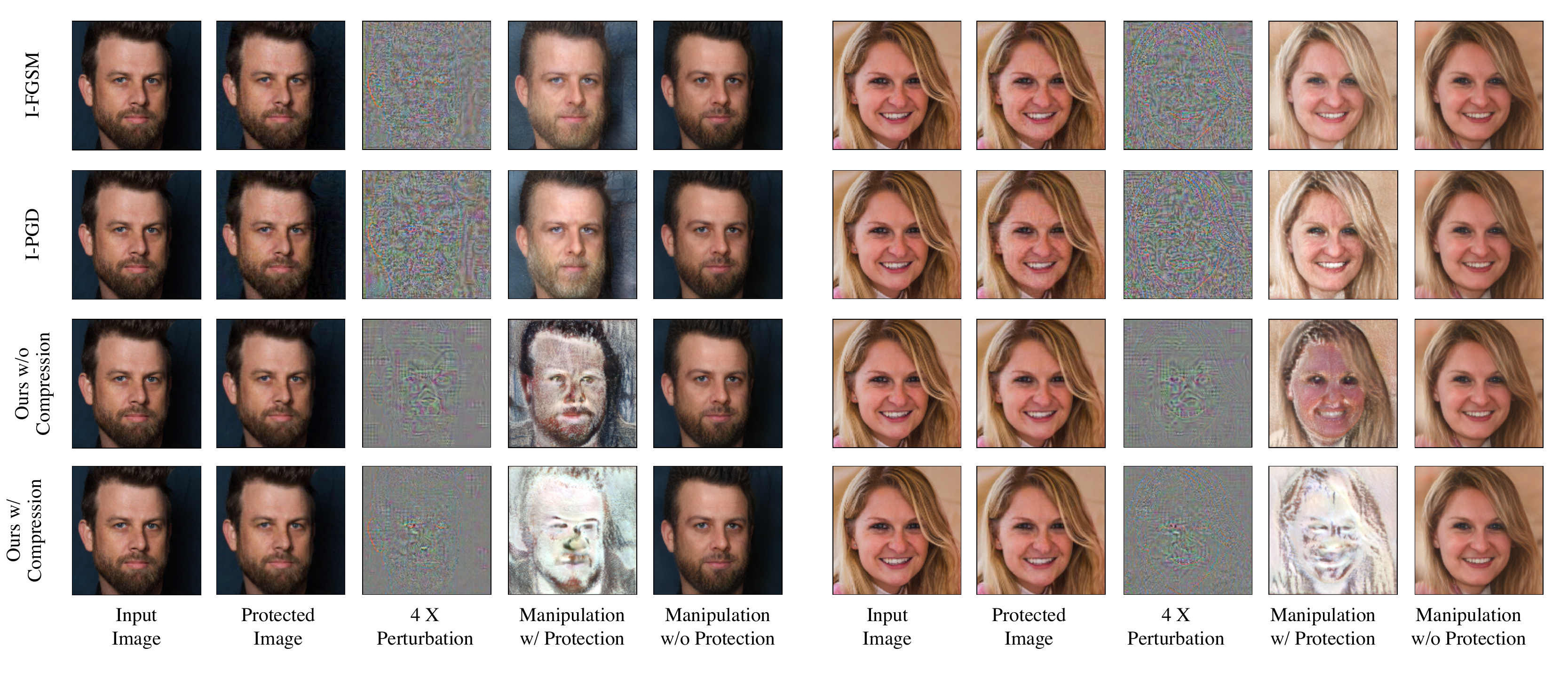}
\end{center}
\vspace{-0.8cm}
  \caption{Qualitative comparison in the presence of JPEG compression (white target). Methods trained without compression struggle; in contrast, our model trained with compression is able to produce perturbations that are robust to compression. \emph{Ours w/ Compression} refers to the model trained with random compression. \emph{Ours w/o Compression} refers to model trained without compression. Compression is applied on the protected images. All methods are evaluated at compression quality C-80.}
\label{fig:compressed_result_self_recon}
\end{figure}

\begin{figure}[h!]
\begin{center}
\includegraphics[width=1.0\linewidth]{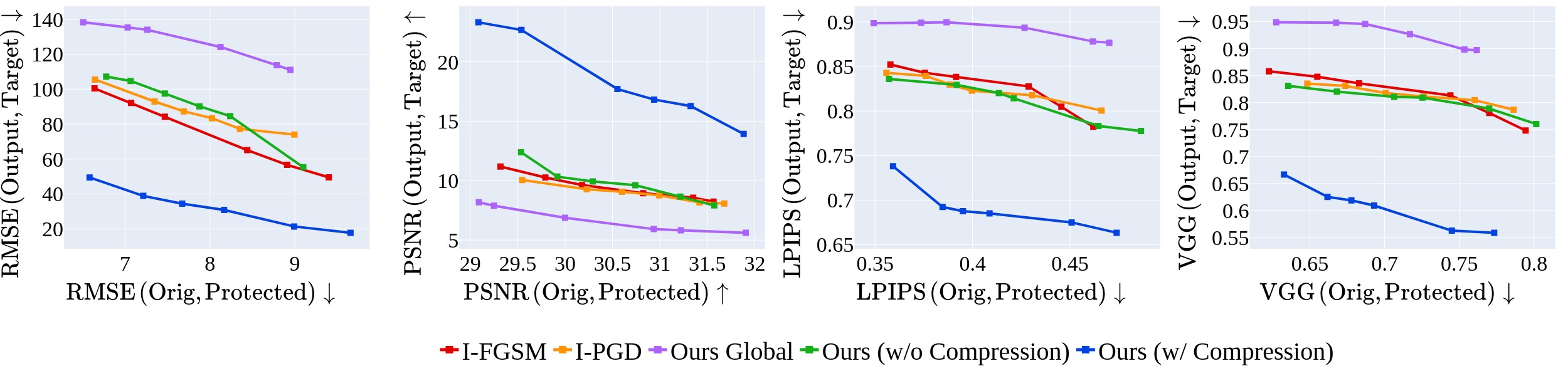}
\end{center}
\vspace{-0.5cm}
   \caption{Performance comparison in the presence of JPEG compression. Our method without differentiable JPEG training manages to disrupt the model; however, training with random compression levels significantly outperforms the uncompressed baselines. All methods are evaluated at compression quality C-80.}
\label{fig:perf_graph_self_recon_compressed}
\end{figure}

\begin{figure}[h!]
\begin{center}
\includegraphics[width=1.0\linewidth]{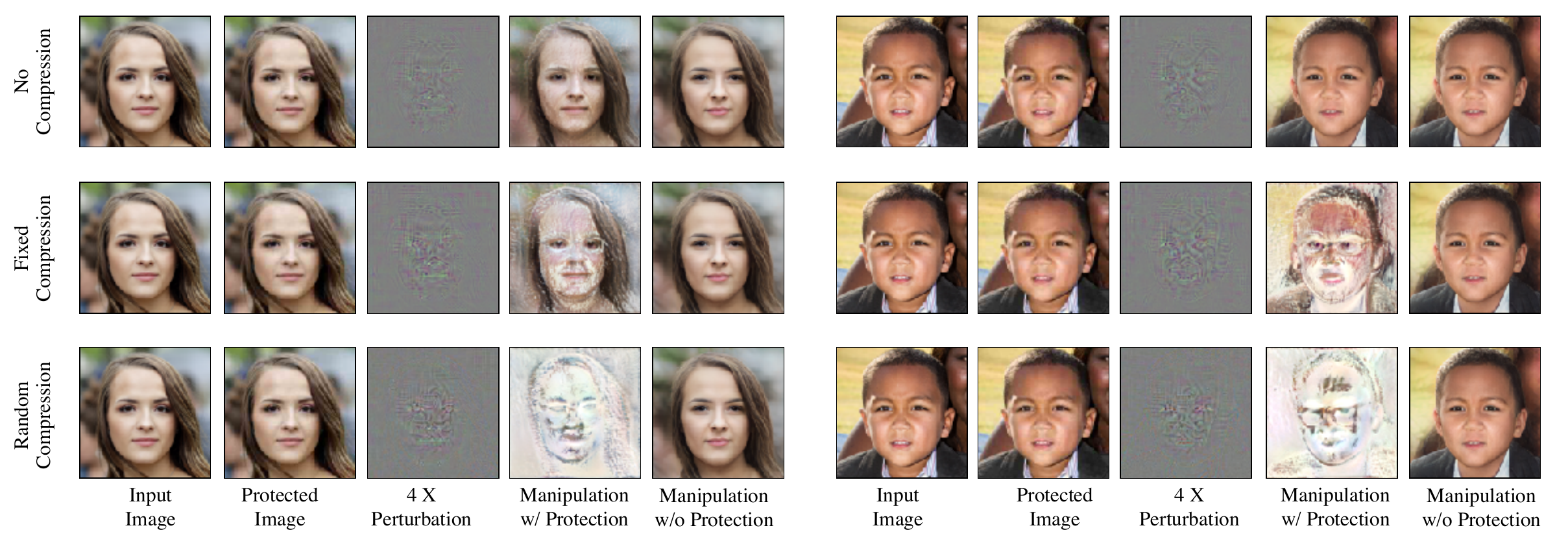}
\end{center}
\vspace{-0.7cm}
  \caption{Comparison for our method trained without compression, fixed compression, and random compression for self-reconstruction task (white target). The fixed compression model was trained with compression quality C-80. All methods are evaluated on compression quality C-30. The randomly compressed model outperforms both fixed and no compression models.}
\label{fig:compressed_result_ablations}
\end{figure}

\textbf{Robustness to JPEG Compression.}
Next, we investigate the sensitivity of perturbations to different compression qualities. 
We apply the actual JPEG compression technique to report results. We observe that without training the model against different compression leads to degraded results when evaluated on compressed images, Fig.~\ref{fig:compressed_result_self_recon} and~\ref{fig:perf_graph_self_recon_compressed}. 
Training the model with fixed compression quality makes the perturbation robust to that specific compression quality; however, it fails for other compression levels; see Fig.~\ref{fig:compressed_result_ablations}. 
We therefore train across different compression levels varied during training iterations.

\begin{figure}[h!]
\begin{center}
\includegraphics[width=1.0\linewidth]{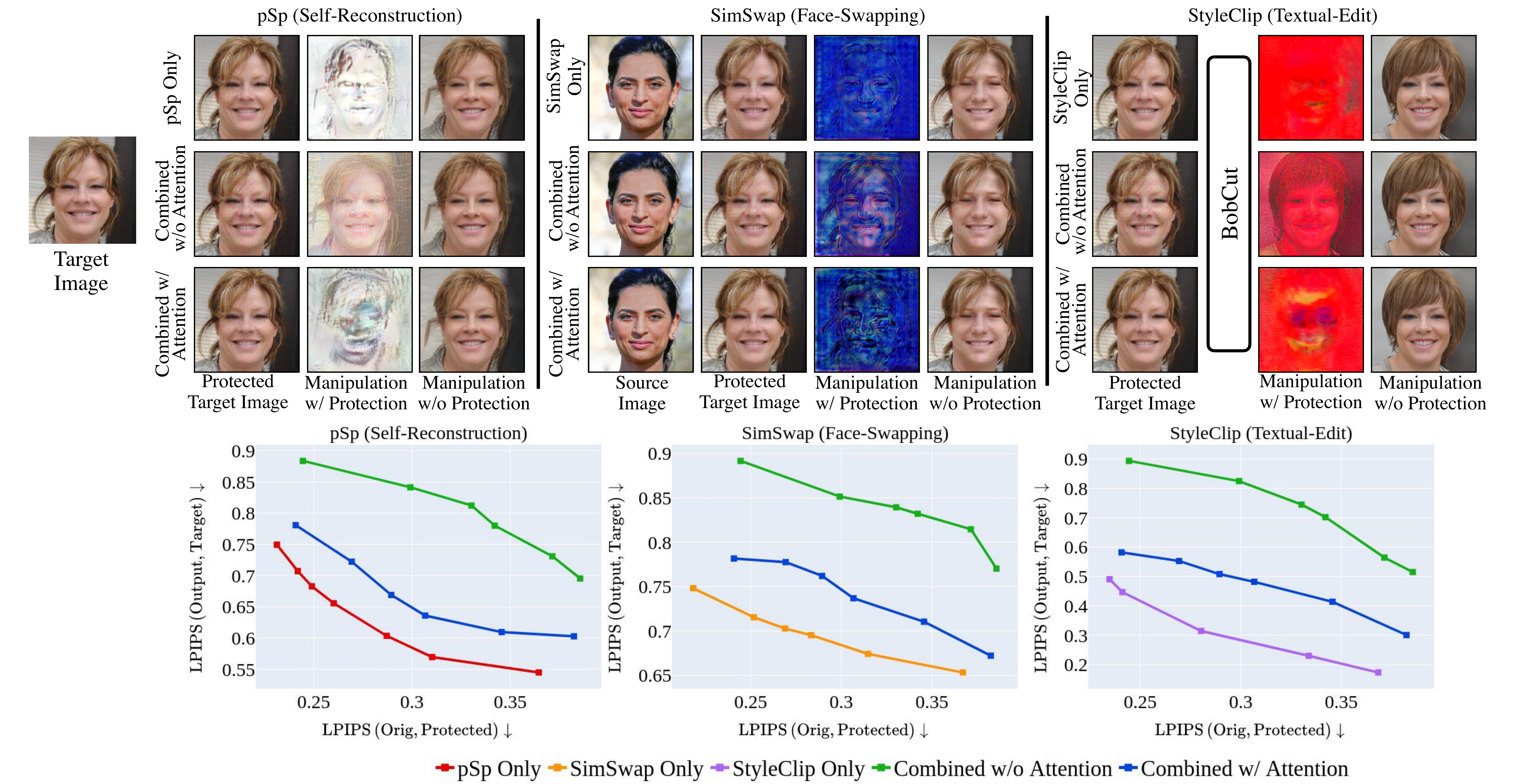}
\end{center}
  \caption{Visual results (top) and performance graph (bottom) for multiple targets simultaneously. \emph{pSp Only}, \emph{SimSwap Only}, and \emph{StyleClip Only} refer to the individual protection models trained only for the respective manipulations. \emph{Combined w/o Attention} refers to a model trained directly for all manipulation methods combined. \emph{Combined w/ Attention} refers to our proposed attention-based fusion approach. Our proposed attention model performs much better the no attention baseline, and is comparable to individual models.}
\label{fig:qualitative_multiple}
\end{figure}

\textbf{Multiple Manipulation Models.}
We leverage manipulation-specific perturbations as priors and combine them using attention-based fusion to generate a single perturbation to protect against multiple manipulations at the same time.
As a baseline, we also compare against a model trained directly for all manipulation methods combined without manipulation-specific priors or attention. We notice that this setup is unable to produce optimal perturbations due to absence of prior information from manipulation-specific perturbations which provide the most optimal perturbations to produce predefined manipulation targets. 
We color-code the manipulation targets with different colors for different manipulation models.
This protection technique has an advantage over simple disruption since it gives more information about which technique was used to manipulate the image. 
We conduct experiments with three different state-of-the-art methods: pSp~\cite{richardson2021encoding} with solid white image as the manipulation target, SimSwap~\cite{ChenCNG20} with solid blue image as target and StyleClip~\cite{Patashnik2021ICCV} with solid red as target image. 
Visual results and performance graph comparison are shown in Fig.~\ref{fig:qualitative_multiple}.
Our combined model with attention produces more effective results than without attention baseline, and without any significant degradation compared to manipulation-specific baselines when handling multiple manipulation at the same time.

\section{Conclusion}

In this work, we proposed a data-driven approach to protect face images from potential popular manipulations. 
Our method can both prevent and simultaneously identify the manipulation technique by generating the predefined manipulation target as output. 
In comparison to existing works, our method not only run orders of magnitude faster, but also achieves superior performance; i.e., with smaller perturbations of a given input image, we can achieve larger disruptions in the respective manipulation methods.
In addition, we proposed an end-to-end compression formulation to make the perturbation robust to compression. 
Furthermore, we propose a new attention-based fusion approach to handle multiple manipulations simultaneously.
We believe our generalized, data-driven method takes an important step towards addressing the potential misuse of popular face image manipulation techniques.

\section*{Acknowledgments}
{
This work is supported by a TUM-IAS Rudolf M\"o{\ss}bauer Fellowship, the ERC Starting Grant Scan2CAD (804724), and Sony Semiconductor Solutions Corporation. We would also like to thank Angela Dai for video voice over.
}
\clearpage
\section*{Supplemental Material}

In this supplemental document, we provide additional details. In the main paper, we demonstrate the effectiveness of our approach on popular face manipulation methods. The motivation behind choosing these specific manipulations is explained in Section~\ref{sec:white_box}. Current forensic methods aim to only identify the manipulated images, which suffers from the limitation of not being able to prevent malicious activities. However, it is practically more relevant to prevent creation of fake images using real images. We achieve this using targeted adversarial attacks, this is explained in Section~\ref{sec:def}.
In Section~\ref{sec:arch}, we provide details about the network architecture and additional training details. We then elaborate the implementation of the baseline methods in Section~\ref{sec:baselines}. Finally, in Section~\ref{sec:add_expts}, we perform ablations and additional experiments.

\begin{figure}[h!]
\begin{center}
\includegraphics[width=0.75\linewidth]{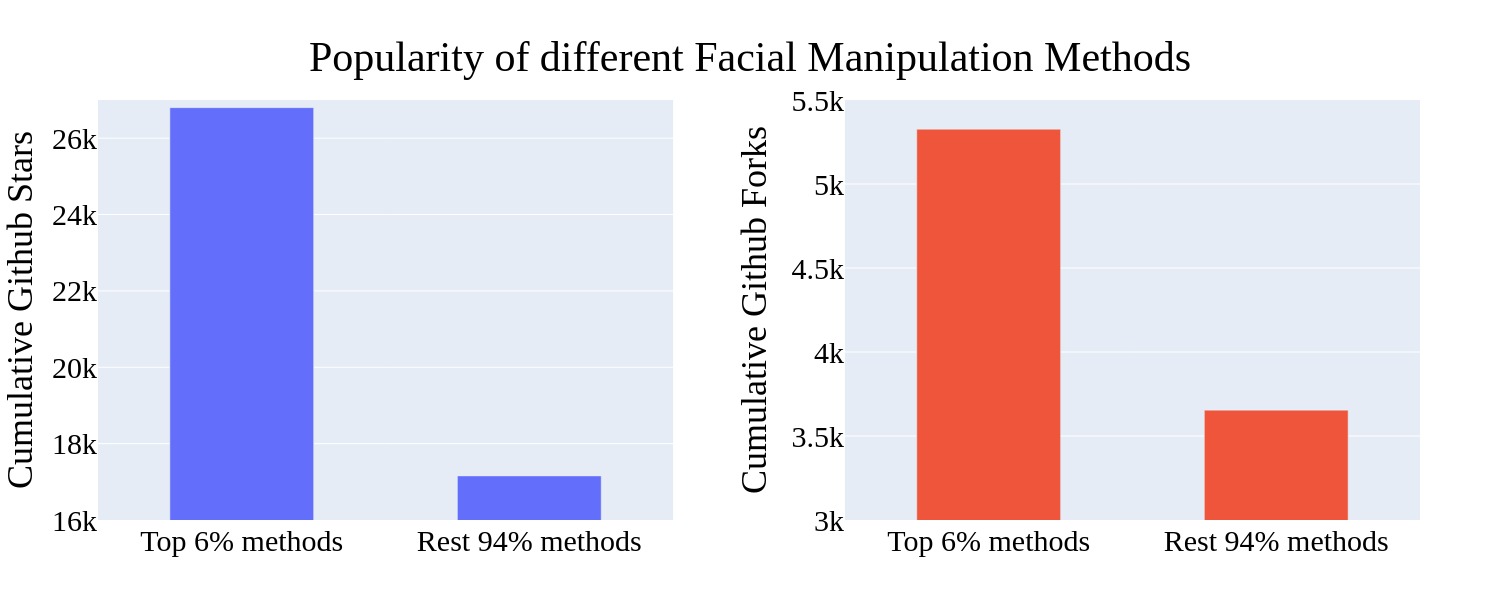}
\end{center}
   \caption{Official github popularity distribution for open-source facial manipulation tools. Methods chosen from last five years' papers from top vision conferences. Note that only a handful of methods show wide applicability in a practical scenario (1.5K+ stars and 300+ forks per repo ) and roughly 94\% of all methods are not used very often.}
\label{fig:github_repos}
\end{figure}
\section{Protection from popular Image Manipulations}\label{sec:white_box}
We emphasize that it is important to protect images from the popular facial manipulations. Compared to large number of open-source image manipulation tools available online, we notice that very few methods have achieved wide applicability among users as shown in Fig.~\ref{fig:github_repos}, thus attracting many users to employ only these popular methods. Secondly, re-training or even running inference on these methods requires specialized knowledge and skills, resulting in common users to look for pre-trained models in an easily accessible way. Therefore, we benchmark our results on these famous high quality facial manipulations that can easily be accessed via web demos, see Tab.~\ref{tab:method_runs}.  We additionally show in Subsection~\ref{sec:transfer_unseen} that our generated perturbations can also protect against new and unseen manipulations built upon these protected manipulations. A complete list of manipulation methods can be found here\footnote{\url{https://bit.ly/35VSd1K}}.

\begin{table}
\begin{center}
\begin{tabular}{ p{2cm}  p{2.0cm} p{2.0cm}  p{4cm} p{2cm}}
\hline\noalign{\smallskip}
\small{Method} & \small{Edit Type} & \small{\# Runs} & \small{URL} & \small{GPU} \\
 &  & \small{(Web Demo)} &  & \small{Hours} \\
\noalign{\smallskip}
\hline
\noalign{\smallskip}
\small{pSp~\cite{richardson2021encoding}}  & \small{Style Mixing} &  \small{5,704} & \small{\url{https://bit.ly/3I7ew22}} & \small{100}\\
\hline
\small{SimSwap~\cite{ChenCNG20}}  & \small{FaceSwap} &  \small{30,944}  & \small{\url{https://bit.ly/3i5mjmh}} & \small{120}\\
\hline
\small{StyleClip~\cite{Patashnik2021ICCV}} & \small{Text-Driven}  &  \small{374,636} & \small{\url{https://bit.ly/3MIShmx}} & \small{12 per style}\\
 &  \small{Manipulation}  &   &  & \small{}\\
\hline
\end{tabular}
\caption{Training time and number of runs (as of 14.03.2022) on the web demo of publicly accessible pre-trained models for the widely used manipulation methods.}
\label{tab:method_runs}
\end{center}
\end{table}

\section{Definitions}\label{sec:def}

\subsection{Targeted Adversarial Attack}
Adversarial attacks~\cite{targeted_attack} are typically used in the context of classification tasks as a technique to fool the classifier model by maximizing the probability of outputting an incorrect target class label other than its actual class.
In a similar spirit, we formalize targeted attacks for face image manipulation as a technique to trick these generative models;
however, in contrast to misleading a classifier, our goal is to produce a predefined target image (uniformly colored solid white/blue/red images in our experiments). 

Recently, Ruiz et al.~\cite{ruiz2020disrupting} proposed the term \textit{targeted disruption}. 
Their key aspect is to use a specific target image to drive the optimization such that it destroys the output of the generative model. 
We on the other side propose to learn a perturbation that forces the manipulation model to produce a specific target image instead of a random output image; we refer to this as a {\em targeted} adversarial attack.

\section{Training Details}\label{sec:arch}

\textbf{Dataset.}
To train our models we define custom split on high resolution FFHQ~\cite{karras2019stylebased} dataset. We additionally evaluate the performance of our models on unseen Celeb-HQ~\cite{karras2018progressive} and VGGFace2-HQ~\cite{github_vgghq} datasets in the main paper. We used 10K images for training and 1000 images for validation and test split each, see Tab.~\ref{tab:data_split} for more details.

\begin{table}[h!]
\begin{center}
\label{table:headings}
  \begin{tabularx}{\textwidth}{c c c c c}
  \toprule
 \textbf{Task} & \textbf{Model} & \textbf{  Mode} &  \textbf{  No. of Images} & \textbf{Dataset} \\ 
\midrule
     \multirow{2}{*}{Self-Reconstruction} & \multirow{2}{*}{pSp~\cite{richardson2021encoding}} & {Train} & {5000} & {FFHQ~\cite{karras2019stylebased}} \\
      &  & {Val} & {1000} & {FFHQ~\cite{karras2019stylebased} }  \\
     \midrule
     \multirow{2}{*}{Face-Swapping} & \multirow{2}{*}{SimSwap~\cite{ChenCNG20}} & {Train} & {2 $\times$ 5000} & {FFHQ~\cite{karras2019stylebased}} \\
      & & {Val} & {2 $\times$ 1000} & {FFHQ~\cite{karras2019stylebased} }  \\
     \midrule
     \multirow{2}{*}{Textual-Editing} & \multirow{2}{*}{StyleClip~\cite{Patashnik2021ICCV}} & {Train} & {5000} & {FFHQ~\cite{karras2019stylebased}} \\
     & &{Val} & {1000} & {FFHQ~\cite{karras2019stylebased} }  \\
     \midrule
     {Self-Reconstruction} & {pSp~\cite{richardson2021encoding}} & {Test} & {1000} & {FFHQ~\cite{karras2019stylebased}}  \\
     \midrule
     {Style-Mixing} & {pSp~\cite{richardson2021encoding}} & {Test} & {2 $\times$ 500} & {FFHQ~\cite{karras2019stylebased}}  \\
     \midrule
     {Face-Swapping} & {SimSwap~\cite{ChenCNG20}} & {Test} & {2 $\times$ 500} & {FFHQ~\cite{karras2019stylebased}}  \\
     \midrule
     {Textual-Editing} & {StyleClip~\cite{Patashnik2021ICCV}} & {Test} & {1000} & {FFHQ~\cite{karras2019stylebased} }\\
     \midrule
     {Style-Mixing} & \multirow{2}{*}{pSp~\cite{richardson2021encoding}} & {Test} & {2 $\times$ 500} & {Celeb-HQ~\cite{karras2018progressive}}  \\
     {(Unseen Dataset)} &  & {Test} & {2 $\times$ 500} & {VGGFace2-HQ~\cite{github_vgghq}} \\
    \bottomrule
  \end{tabularx}
  \caption{Dataset split for different tasks used in our experiments. For self-reconstruction, the manipulation model takes one image as input. For style-mixing and face-swapping, two images are fed as input to the manipulation model. For textual-editing, an image and a text prompt describing the manipulation-style is given an input to the model. For consistency, all methods are benchmarked on the same set of images (except unseen images).}
  \label{tab:data_split}
\end{center}
\end{table}

\begin{figure*}[h!]
\begin{center}
\includegraphics[width=\linewidth]{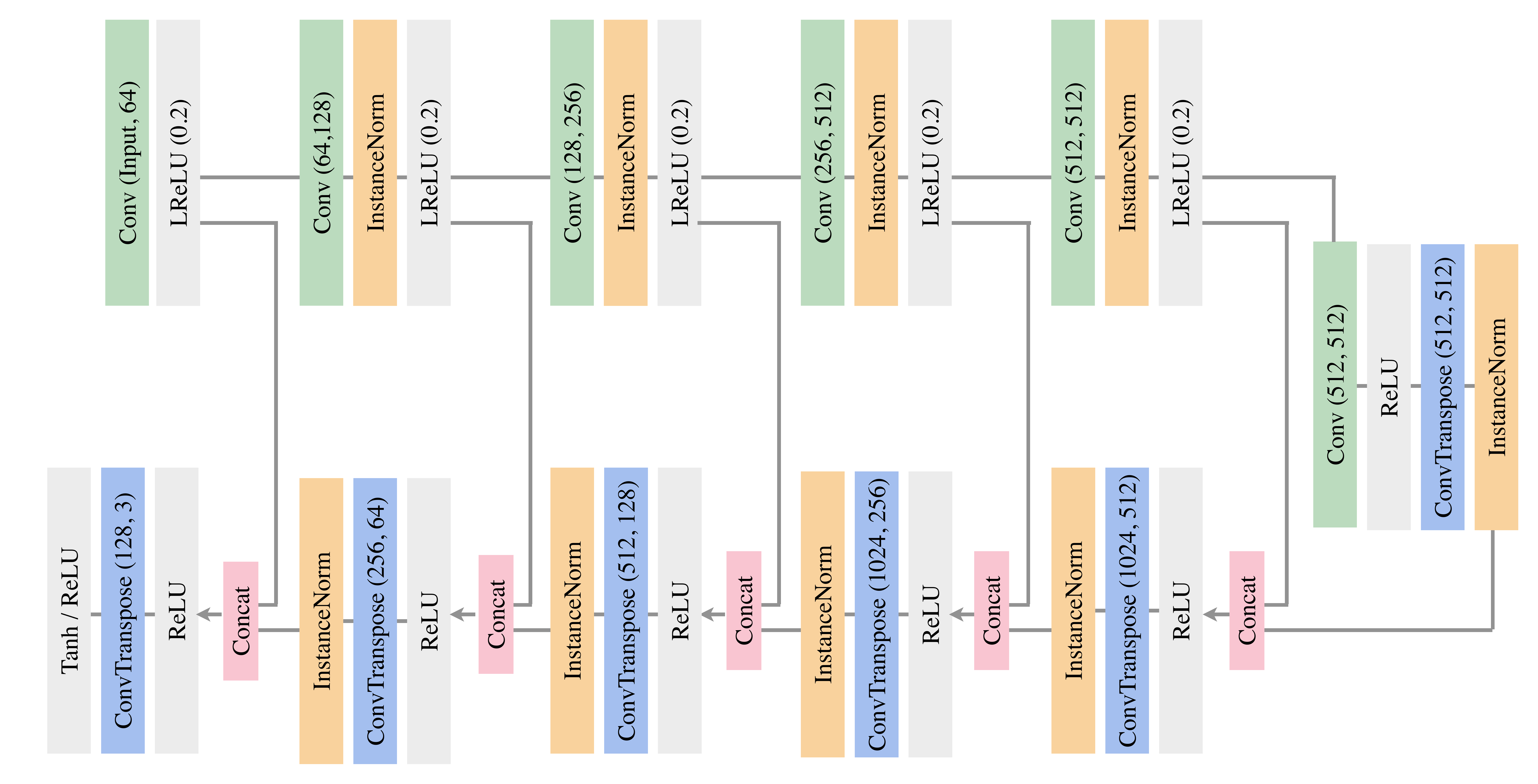}
\end{center}
\caption{Architecture Overview: Network architecture used for experiments. Conv(.) and ConvTranspose(.) refer to the 2D convolution and 2D transposed convolution operation. For both Conv and ConvTranspose, we use a kernel size of $4 \times 4$. For conv layers, values in bracket refers to the input and output channels; e.g., Conv($X$, $Y$) denotes a $X$ channel input and $Y$ channel output. The input channel for the first Conv varies for different backbones: 6 for ProtectionNet $\boldsymbol{g_\Phi}$, 4 for AttentionNet $\boldsymbol{h_\omega}$ and 3 for FusionNet $\boldsymbol{r_\rho}$. LReLU(0.2) denotes the LeakyReLU activation with a negative slope 0.2. Concat denotes the concatenation operation. For ProtectionNet $\boldsymbol{g_\Phi}$ and FusionNet $\boldsymbol{r_\rho}$, we use Tanh non-linearity after the last layer and ReLU for AttentionNet $\boldsymbol{h_\omega}$.}
\label{fig:model_arch}
\end{figure*}

\textbf{Model Architecture.}
For all three model backones used in the paper ProtectionNet $\boldsymbol{g_\Phi}$, AttentionNet $\boldsymbol{h_\omega}$, and FusionNet $\boldsymbol{r_\rho}$, we use a convolutional neural network based on U-Net~\cite{ronneberger2015convolutional} architecture. More specifically, we use UNet-64 architecture with 29.24M parameters, as shown in Fig.~\ref{fig:model_arch}. The weights of the network are normal initialized with scaling factor of 0.02. 

For ProtectionNet $\boldsymbol{g_\Phi}$, the network takes a 6 channel input, channel-wise concatenated original image $\textbf{X}_i$ and the globally optimized perturbation $\boldsymbol{\delta}_G$ to predict the image-specific perturbation $\boldsymbol{\delta}_i$. For AttentionNet $\boldsymbol{h_\omega}$, the network takes a 4 channel input, the image-specific perturbation $\boldsymbol{\delta}_i$ and manipulation method label $C_k$ and generates spatial attention map $\alpha_i^k$. For FusionNet $\boldsymbol{r_\rho}$, the model takes a 3 channel input, the blended model-specific perturbation with spatial attention maps and produces model-agnostic perturbation $\boldsymbol{\delta}_i^{\text{all}}$. 
\\
\

\section{Baseline Implementations}\label{sec:baselines}
We compare our method against adversarial attack baselines I-FGSM~\cite{Kurakin2017AdversarialML} and I-PGD~\cite{Madry2018TowardsDL}. 
These methods target classification tasks, where attack patterns are optimized for each image individually. 
To this end, we adapt their method for our targeted adversarial attack task on generative models.
Similar to their original implementations, we define a loss that only consists of the reconstruction term; i.e.,
$$
    \mathcal{L}^{\text{Total}} =   \mathcal{L}^{\text{recon}} = \Big\| \boldsymbol{f_{\Theta}}(\mathbf{X}^{p}) - \mathbf{Y}^{\text{target}} \Big\|_2 ,
$$
where $\mathbf{X}^p$ refers to the protected image and $\mathbf{Y}^{\text{target}}$ refers to the predefined manipulation target (solid color white/blue image in our case).
The amount of perturbation in the image can be controlled with the magnitude of the update step.

\paragraph{I-FGSM}~\cite{Kurakin2017AdversarialML} (Iterative Fast Gradient Sign Method): The protected image is first initialized with the original image as:
$$
    \mathbf{X}_0^{{p}} := \mathbf{X},
$$
and then updated iteratively as
$$
    \mathbf{X}_{n+1}^{{p}} =  \text{Clamp}_{\varepsilon} \left\{ \mathbf{X}_{n}^{{p}} - \alpha \text{ sign}(\nabla_{X}\mathcal{L}^{\text{recon}}(\mathbf{X}_{n}^{{p}}, \mathbf{Y}^{\text{target}}))  \right\} ,
$$
where $\alpha$ is the perturbation strength, $\text{Clamp}_{\varepsilon}(\xi)$ clips the higher and lower values in the valid range $[-\epsilon, \epsilon]$ and $\text{sign}(\zeta)$ returns the sign vector of $\zeta$. We report results with 100 iterations.

\paragraph{I-PGD}~\cite{Madry2018TowardsDL} Iterative Projected Gradient Descent):  The protected image is obtained by the following update steps:
$$
    \mathbf{X}_0^{{p}} := \mathbf{X},
$$
$$
    \mathbf{X}_{n+1}^{{p}} =  \Pi_{\mathcal{S}} \left\{ \mathbf{X}_{n}^{{p}} - \alpha \text{ sign}(\nabla_{X}\mathcal{L}^{\text{recon}}(\mathbf{X}_{n}^{{p}}, \mathbf{Y}^{\text{target}}))  \right\}
$$
where $\alpha$ is the step size and $\Pi_{\mathcal{S}}(.)$ refers to the projection operator that projects onto the feasible set ${\mathcal{S}}$. We report results with 100 iterations and a step size of 0.01.

\section{Additional Experiments}\label{sec:add_expts}
In this section, we report results for some additional experiments. We first show the performance of our model on unseen datasets in Subsection~\ref{sec:unseen_datasets}. Next we compare our method against some additional baselines in Subsection~\ref{sec:additional_baselines}. We then show additional visuals for multiple manipulations simultaneously in Subsection~\ref{sec:multiple_methods}. Next, we analyze the effect of incrementally adding more manipulation methods in Subsection~\ref{sec:more_methods}. We then investigate the importance of AttentionNet backbone for blending manipulation-specific perturbations in Subsection~\ref{sec:attn_backbone}. We also show that our generated perturbations can efficiently protect against related unseen manipulations in Subsection~\ref{sec:transfer_unseen}.  We then analyze the distribution of different baselines in Subsection~\ref{sec:anaysis_baselines}. Then, we report the performance of our method for different norms in Subsection~\ref{sec:diff_norms}. In Subsection~\ref{sec:multi_level_comp}, we evaluate the robustness of our method when compression is applied multiple times to the image. We show visual results for high perturbation levels in Subsection~\ref{sec:high_comp}. Finally, in Subsection~\ref{sec:diff_jpeg_approxs}, we compare different $\round$ approximations for quantization step for the differentiable JPEG compression implementation. For brevity, we show results on self-reconstruction task with white image as manipulation target, unless otherwise stated.

\subsection{Unseen Datasets}\label{sec:unseen_datasets}
\begin{figure}
\begin{center}
\includegraphics[width=1.0\linewidth]{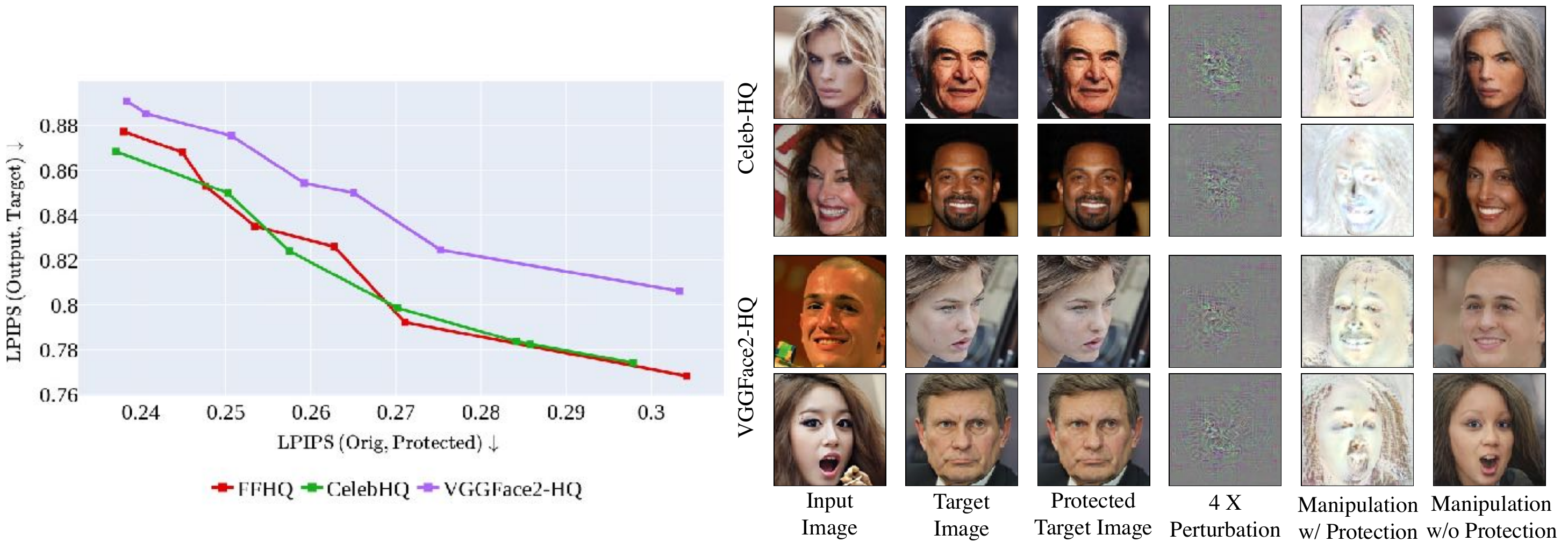}
\end{center}
  \caption{Performance of unseen datasets on the style-mixing task with white manipulation target. The models are trained on FFHQ~\cite{karras2019stylebased} and evaluated on unseen Celeb-HQ~\cite{karras2018progressive} and VGGFace2-HQ~\cite{github_vgghq} datasets. The protection is applied to the target image. Perturbation is enlarged by a factor of four for better visibility. Our model demonstrates comparable performance on these unseen datasets compared to the seen FFHQ dataset~\cite{karras2019stylebased}. }
\label{fig:uncompressed_results_unseen_dataset}
\end{figure}

We further compare the performance of our method on two high-resolution unseen datasets, Celeb-HQ~\cite{karras2018progressive} and VGGFace2-HQ~\cite{github_vgghq}. These datsets were neither seen during training or validation. We simply evaluate our trained models on these datasets. We show results on the style mixing task in Fig.~\ref{fig:uncompressed_results_unseen_dataset}. Note that our model demonstrates comparable performance on these unseen datasets compared to the seen FFHQ dataset~\cite{karras2019stylebased}.

\subsection{Comparison with additional baselines}\label{sec:additional_baselines}
\begin{figure}[t!]
\begin{center}
{\includegraphics[width=0.7\linewidth]{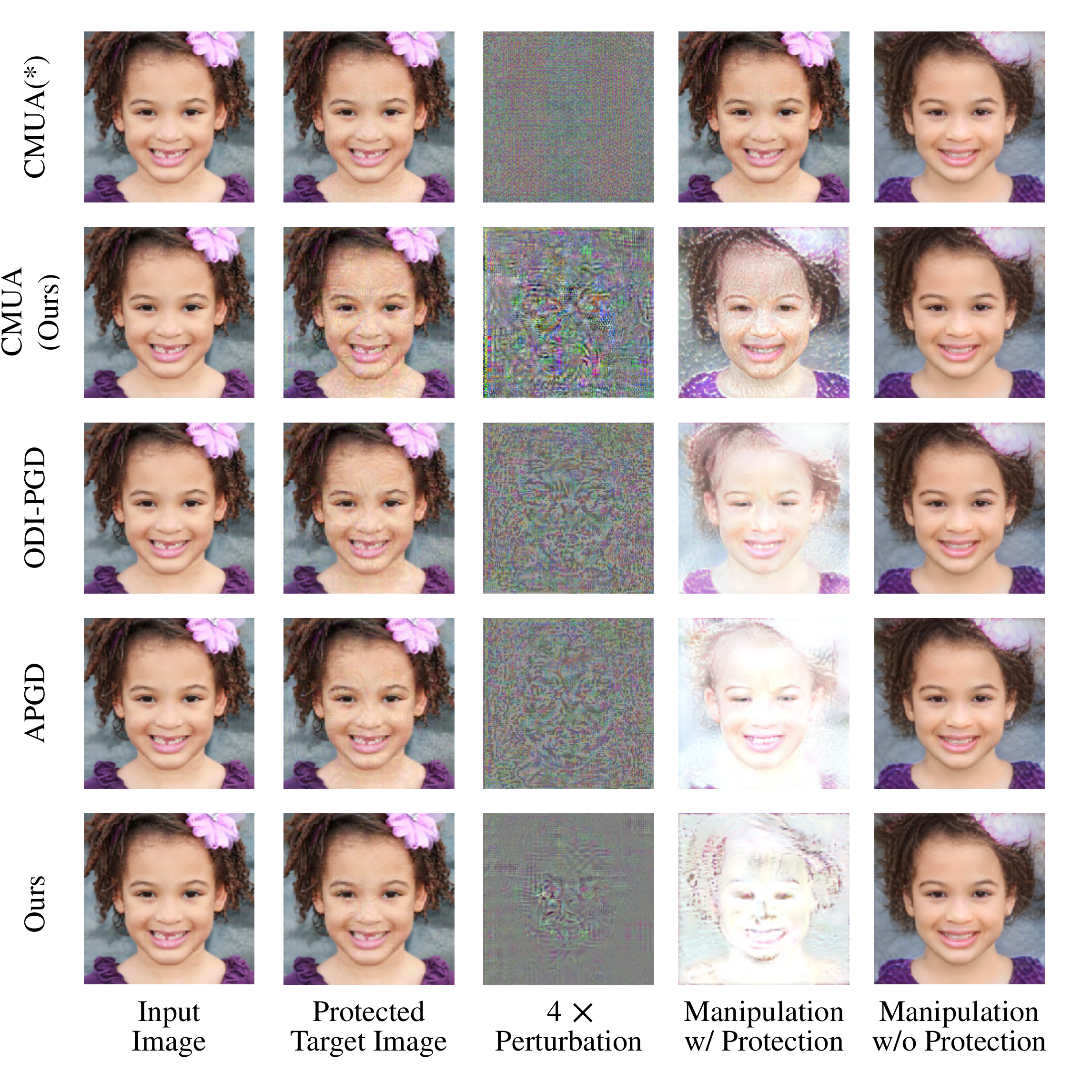} }%
\end{center}
   \caption{\small{Comparison on pSp (white target): CMUA(*) and CMUA (Ours) refers to their pretrained patterns and ours adapted.}}
\label{fig:baselinec_imgs}
\end{figure}

\begin{figure}[h!]
\begin{center}
{\includegraphics[width=0.75\linewidth]{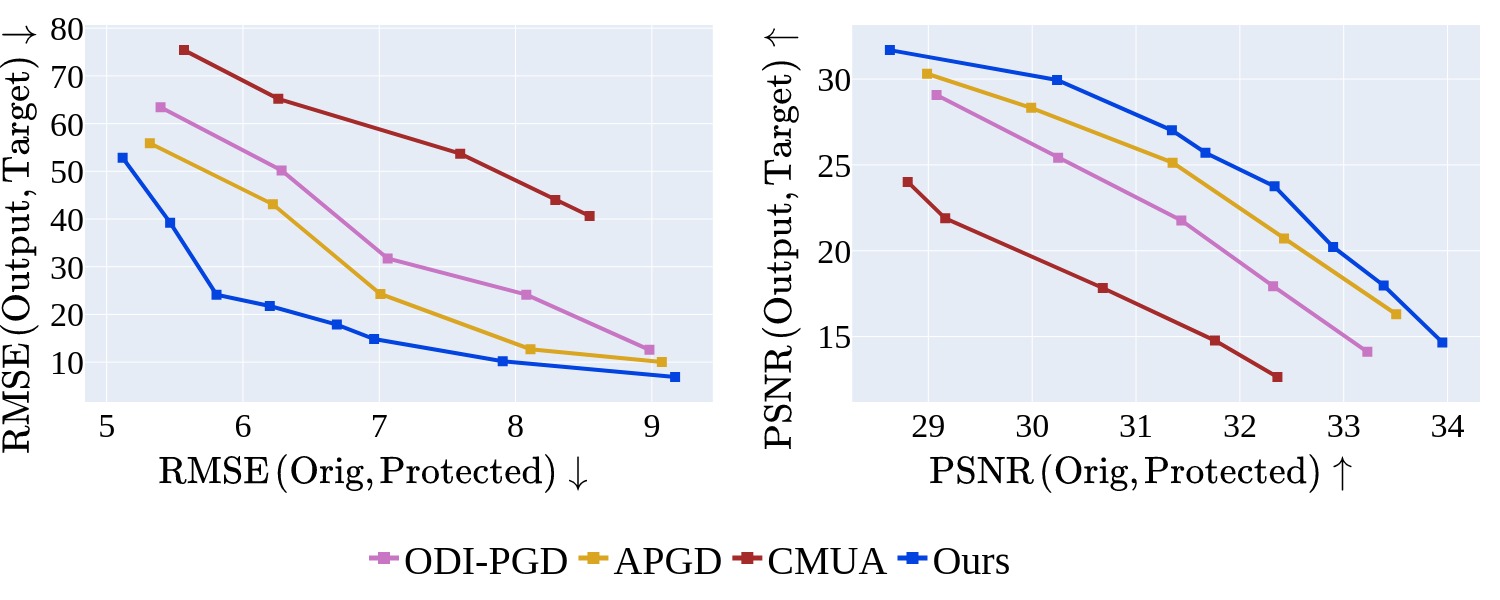} }%
\end{center}
   \caption{\small{Quantitative comparison with baselines on pSp (white target).}}
\label{fig:baselinec_perf}
\end{figure}

Here we compare our method against some additional baseline methods. ODI-PGD~\cite{odi_pgd} (17751.72 ms) and APGD~\cite{apgd} (25786.96 ms) perform per-image optimization, and are therefore significantly slower; CMUA~\cite{huang2022cmua} is faster during inference; however, it produces lower-quality targeted disruptions compared to others. We outperform these baselines; see Fig.~\ref{fig:baselinec_imgs} and Fig.~\ref{fig:baselinec_perf}. 

\subsection{Additional Results for Multiple Manipulations Simultaneously}\label{sec:multiple_methods}
\begin{figure}[h!]
\begin{center}
\includegraphics[width=1.0\linewidth]{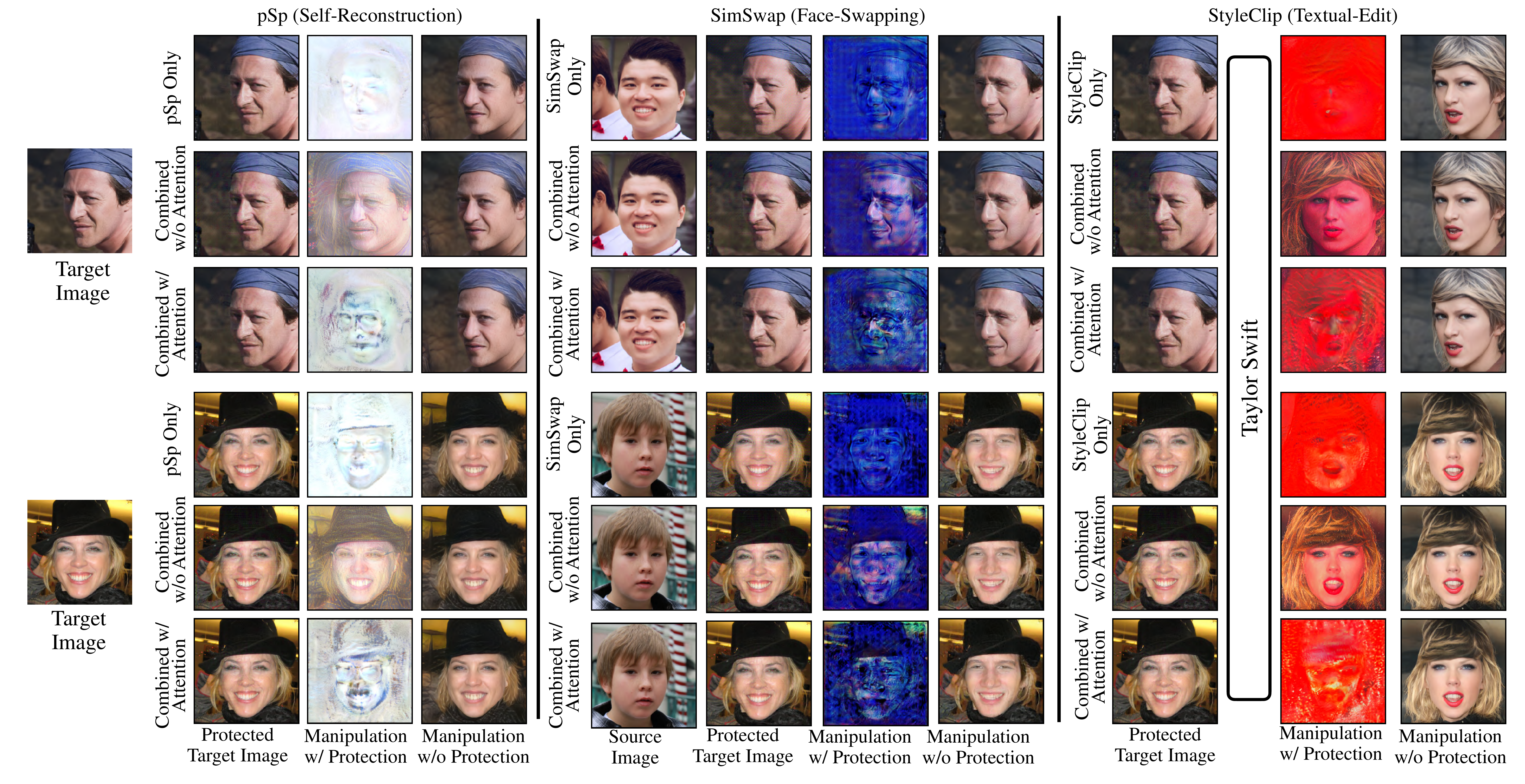}
\end{center}
\caption{Visual results for multiple targets simultaneously. \emph{pSp Only}, \emph{SimSwap Only}, and \emph{StyleClip Only} refer to the individual protection models trained only for the respective manipulations. \emph{Combined w/o Attention} refers to a model trained directly for all manipulation methods combined. \emph{Combined w/ Attention} refers to our proposed attention-based fusion approach. Our proposed attention model performs much better than the no-attention baseline, and is comparable to individual models.}
\label{fig:more_visuals_combined}
\end{figure}

Due to space limitation in the main paper, we show additional visual results for generating manipulation-agnostic perturbations to protect against multiple manipulation methods at the same time, see Fig.~\ref{fig:more_visuals_combined}.

\begin{figure}[h!]
\begin{center}
\includegraphics[width=1.0\linewidth]{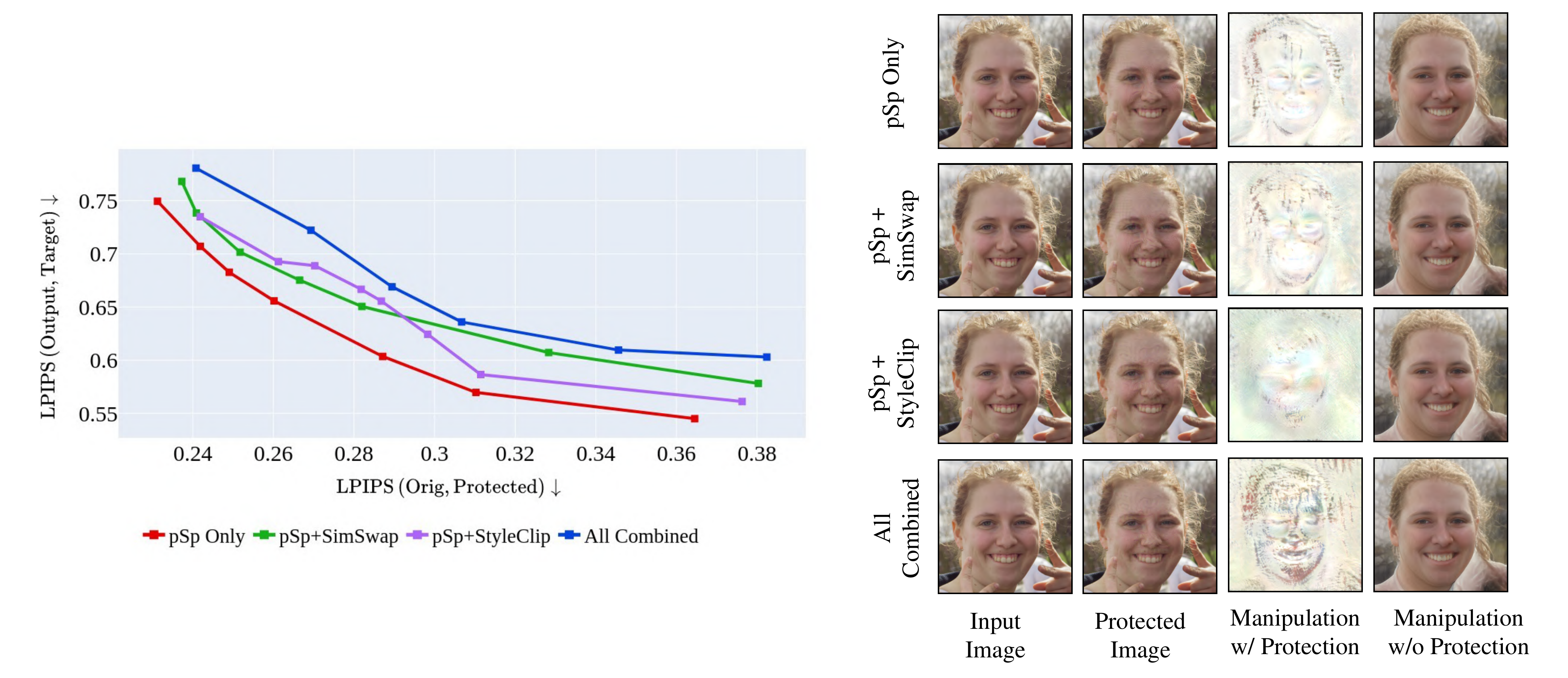}
\end{center}
\caption{Performance graph (left) and visual results (right) for pSp model~\cite{richardson2021encoding} when incrementally adding multiple manipulations. \emph{pSp Only} refers to the protection model trained only for pSp. \emph{pSp + SimSwap} refers to a model trained for pSp~\cite{richardson2021encoding} and SimSwap~\cite{ChenCNG20} manipulations combined. \emph{pSp + StyleClip} refers to a model trained for pSp~\cite{richardson2021encoding} and StyleClip~\cite{Patashnik2021ICCV} manipulations combined. \emph{All Combined} refers to a model trained for all three manipulation methods combined. Our method can be easily extended to handle multiple manipulations. }
\label{fig:more_methods_pSp}
\end{figure}

\begin{figure}[h!]
\begin{center}
\includegraphics[width=1.0\linewidth]{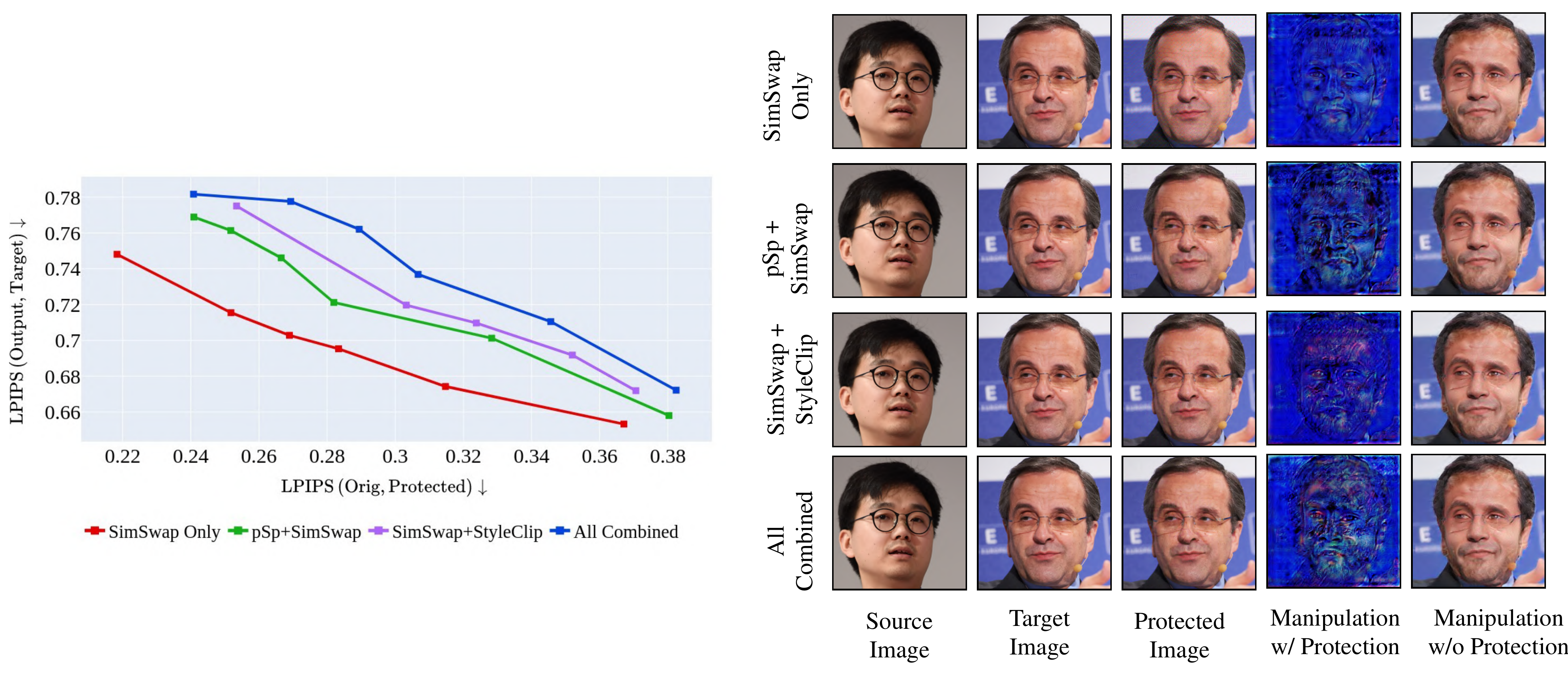}
\end{center}
\caption{Performance graph (left) and visual results (right) for SimSwap~\cite{ChenCNG20} when incrementally adding multiple manipulations. \emph{SimSwap Only} refers to the protection model trained only for SimSwap. \emph{pSp + SimSwap} refers to a model trained for pSp~\cite{richardson2021encoding} and SimSwap~\cite{ChenCNG20} manipulations combined. \emph{SimSwap + StyleClip} refers to a model trained for SimSwap~\cite{ChenCNG20} and StyleClip~\cite{Patashnik2021ICCV} manipulations combined. \emph{All Combined} refers to a model trained for all three manipulation methods combined. Our method can be easily extended to handle multiple manipulations.}
\label{fig:more_methods_simswap}
\end{figure}

\begin{figure}[h!]
\begin{center}
\includegraphics[width=1.0\linewidth]{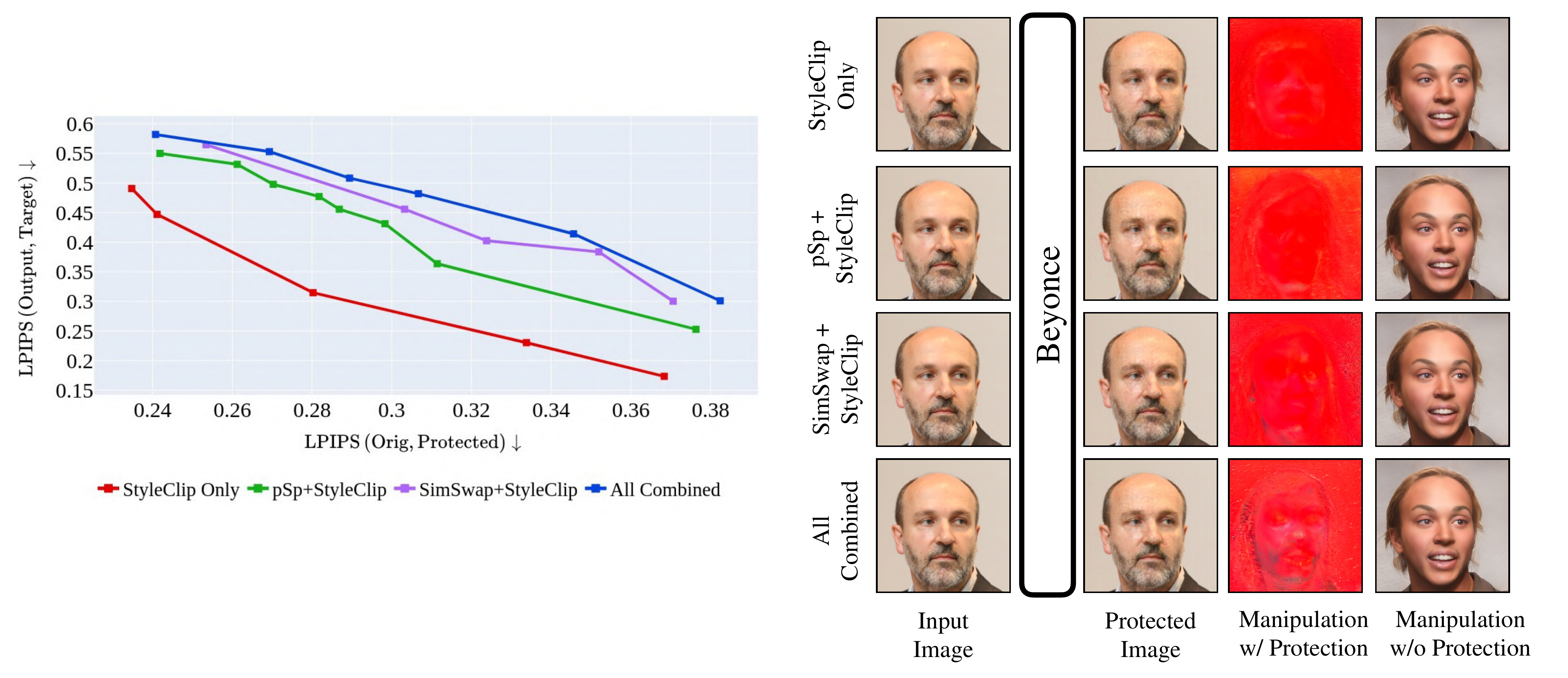}
\end{center}
\caption{Performance graph (left) and visual results (right) for StyleClip~\cite{Patashnik2021ICCV} when incrementally adding multiple manipulations. \emph{StyleClip Only} refers to the protection model trained only for StyleClip. \emph{pSp + StyleClip} refers to a model trained for pSp~\cite{richardson2021encoding} and StyleClip~\cite{Patashnik2021ICCV} manipulations combined. \emph{SimSwap + StyleClip} refers to a model trained for SimSwap~\cite{ChenCNG20} and StyleClip~\cite{Patashnik2021ICCV} manipulations combined. \emph{All Combined} refers to a model trained for all three manipulation methods combined. Our method can be easily extended to handle multiple manipulations.}
\label{fig:more_methods_styleclip}
\end{figure}

\subsection{Effect of adding more manipulation methods}\label{sec:more_methods}
We additionally analyze the effect of incrementally adding more manipulation methods using our proposed attention mechanism. We perform experiments with different combinations of manipulation methods. Results for pSp Encoder~\cite{richardson2021encoding}, SimSwap~\cite{ChenCNG20} and StyleClip~\cite{Patashnik2021ICCV} are shown in Fig.~\ref{fig:more_methods_pSp},~\ref{fig:more_methods_simswap} and~\ref{fig:more_methods_styleclip} respectively. Note that gradually adding more methods using our proposed attention mechanism can be easily extended to handle multiple methods at the same time. This indicates that new manipulations can be easily integrated without significant deterioration in the overall performance on the individual manipulation(s).

\subsection{Analysis of AttentionNet Backbone}\label{sec:attn_backbone}
\begin{figure}[h!]
\begin{center}
\includegraphics[width=1.0\linewidth]{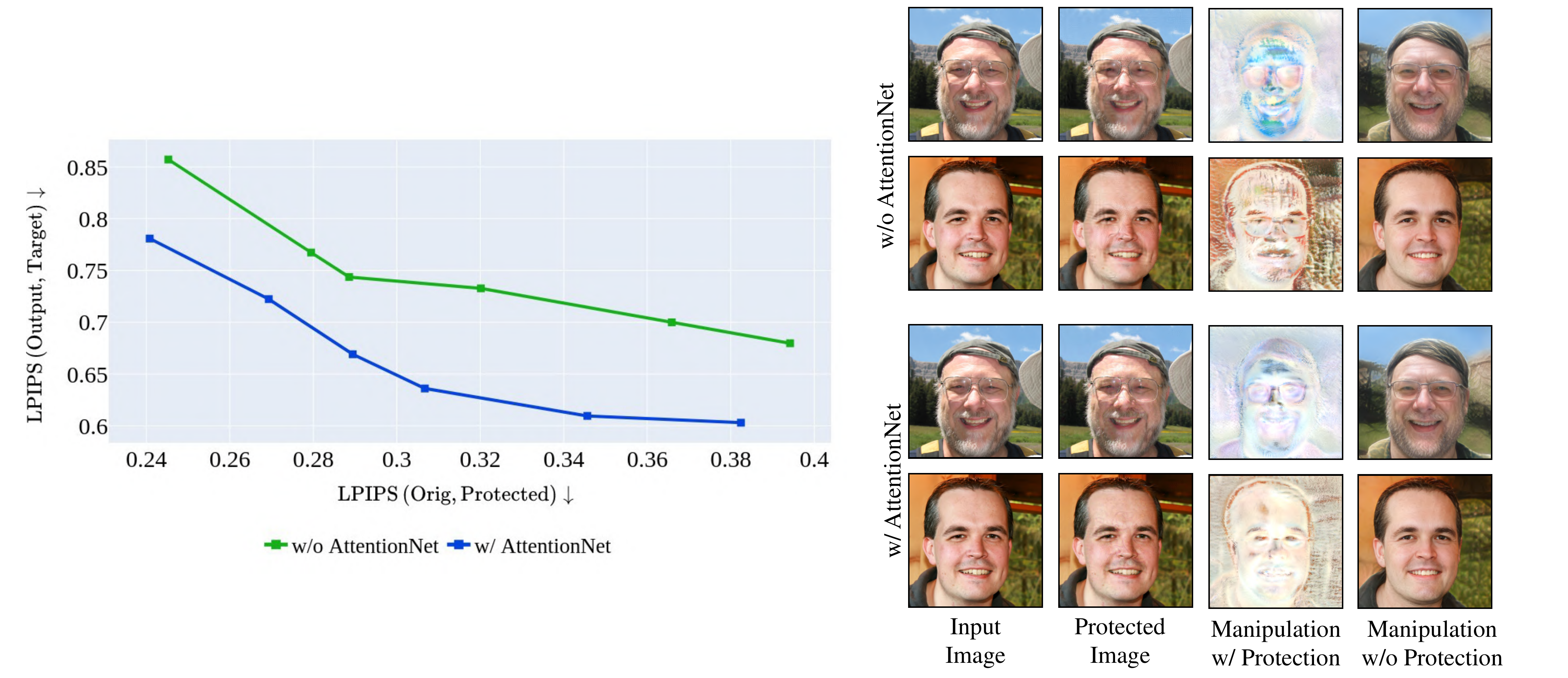}
\end{center}
\caption{Analysis of AttentionNet backbone. \emph{w/o AttentionNet} refers to direct blending of manipulation-specific perturbations using FusionNet only, without attention. \emph{w/ AttentionNet} refers to blending of manipulation-specific perturbations using both AttentionNet and FusionNet. \emph{w/ AttentionNet} outperforms \emph{w/o AttentionNet} baseline indicating that using attention efficiently blends manipulation-specific perturbations.}
\label{fig:attention}
\end{figure}

Next, we analyze the importance of attention network backbone to produce manipulation-agnostic perturbation when handling multiple manipulations at the same time. We notice that directly fusing the manipulation-specific perturbations by only using the FusionNet backbone leads to suboptimal results. Thus, the attention backbone serves as an important component to generate efficient patterns for all methods together. Results for this experiment are shown in Fig.~\ref{fig:attention}. 
Note that this experiment differs from \textit{Combined w/o Attention} baseline comparison shown in the main paper, which refers to directly learning a single model for multiple manipulations combined (without AttentionNet and FusionNet backbone). Here we compare blending of manipulation-specific perturbations using FusionNet only (without AttentionNet).

 \begin{table}
\begin{center}
\begin{tabular}{ p{4cm} p{2.0cm}  p{5cm}}
\hline\noalign{\smallskip}
\small{Method} & \small{\# Runs} & \small{URL (Web Demo)} \\
\noalign{\smallskip}
\hline
\noalign{\smallskip}
\small{SAM~\cite{alaluf2021matter}} &  \small{18,205} & \small{\url{https://bit.ly/3KG7F19}} \\
\hline
\small{Style-NADA~\cite{gal2021stylegannada}}  &  \small{22,031}  & \small{\url{https://bit.ly/3J4Da4F}} \\
\hline
\end{tabular}
\vspace{0.2cm}
\caption{Number of runs on the web demo (as of 14.03.2022) of publicly accessible pre-trained models for the unseen manipulation methods.}
\label{tab:method_runs_unseen}
\end{center}
\end{table}

\begin{figure}[h!]
\begin{center}
\includegraphics[width=1.0\linewidth]{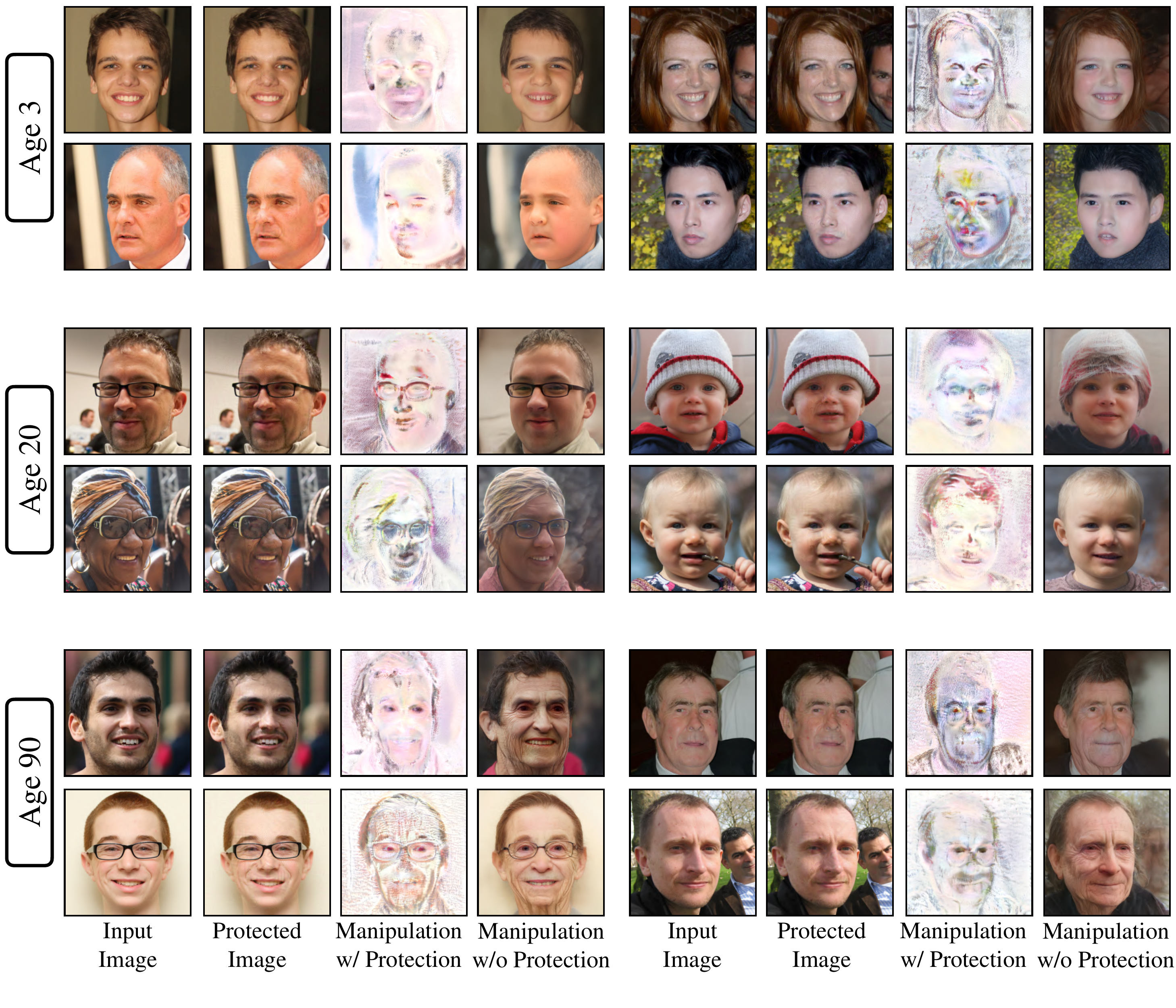}
\end{center}
\caption{Protection results on unseen Age Transformation Model SAM~\cite{alaluf2021matter}. Our model trained on a related method pSp~\cite{richardson2021encoding} can protect against this new age transformation manipulation. Note that our protection is robust to different age transformations.}
\label{fig:age}
\end{figure}

\begin{figure}[h!]
\begin{center}
\includegraphics[width=1.0\linewidth]{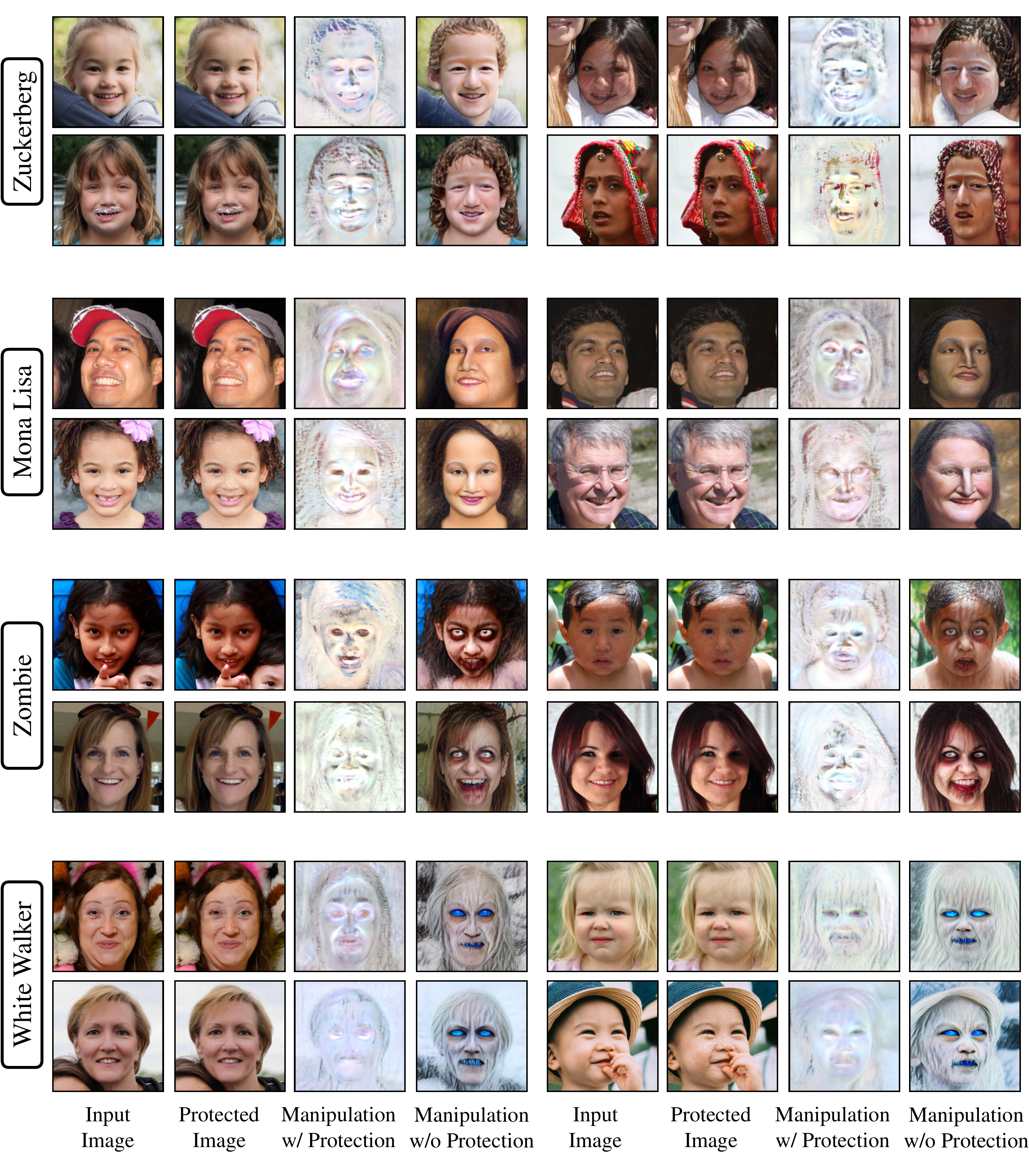}
\end{center}
\caption{Protection results on unseen models from Style-NADA~\cite{gal2021stylegannada} with different manipulation types. Our model trained on a related method pSp~\cite{richardson2021encoding} can protect against these new image manipulations. Our generated perturbation can protect against all the different variants of manipulations generated by different Style-NADA models.}
\label{fig:nada}
\end{figure}

\subsection{Transferability on Related Models}\label{sec:transfer_unseen}
We further show that our perturbations can even provide protection against unseen related methods build using existing methods. We show results on two different methods: (1) SAM~\cite{alaluf2021matter}, an age transformation model. (2) Style-NADA~\cite{gal2021stylegannada}, a domain adaptation technique for image generators using only text prompt. Both of these methods are built upon pSp~\cite{richardson2021encoding} to encoder images, which is protected with white target in our experiments. For SAM~\cite{alaluf2021matter}, we show results across different age distributions. For Style-NADA~\cite{gal2021stylegannada}, we show results on several different manipulation models/styles. Visual results are shown in Fig.~\ref{fig:age} and Fig.~\ref{fig:nada} and details on web demo are provided in Tab.~\ref{tab:method_runs_unseen}. 

Our applied protection can efficiently protect against both these unseen manipulations indicating that our generated perturbations can defend from newer methods build upon existing protected methods. Note that we do not train our model to protect against these new methods SAM~\cite{alaluf2021matter} and Style-NADA~\cite{gal2021stylegannada}. We only evaluate the robustness of our generated perturbations on these models.   

\subsection{Analysis of different baselines}\label{sec:anaysis_baselines}
\begin{figure}[h!]
\begin{center}
\includegraphics[width=0.7\linewidth]{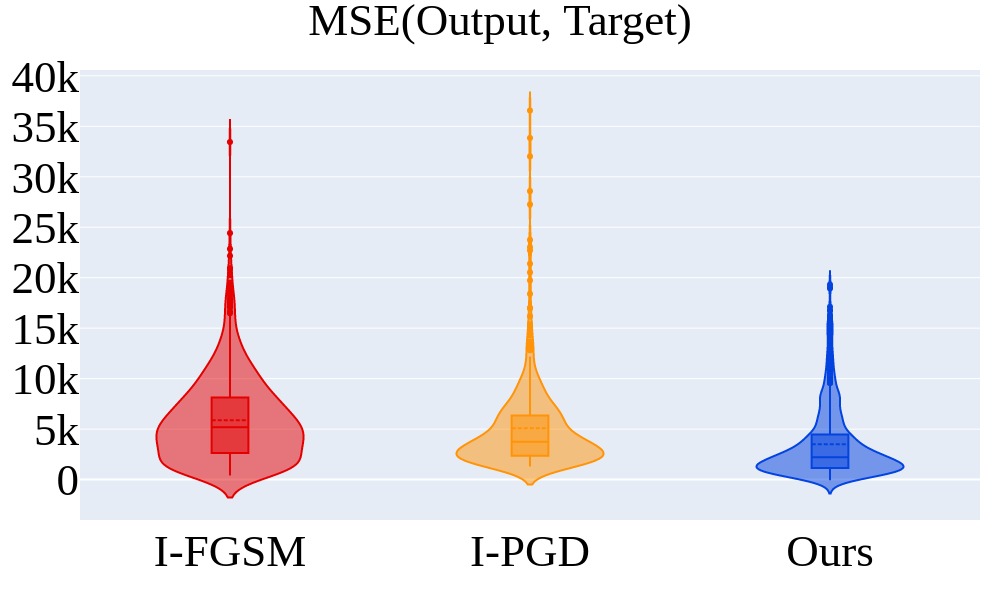}
\end{center}
\caption{Results for the self-reconstruction task~\cite{richardson2021encoding} with white image as manipulation target. We show violin plots for different methods visualizing the mean squared error of the manipulation model output with the predefined manipulation target image. Our method, denoted as Ours, outperforms alternate methods showing a lot less variance in the output distribution. In contrast, per-image optimization techniques such as I-PGD have long tail distribution indicating that the method is not equally effective for all the samples in the dataset.}
\label{fig:self_recon_violin_plot}
\end{figure}

\begin{figure}[h!]
\begin{center}
\includegraphics[width=0.7\linewidth]{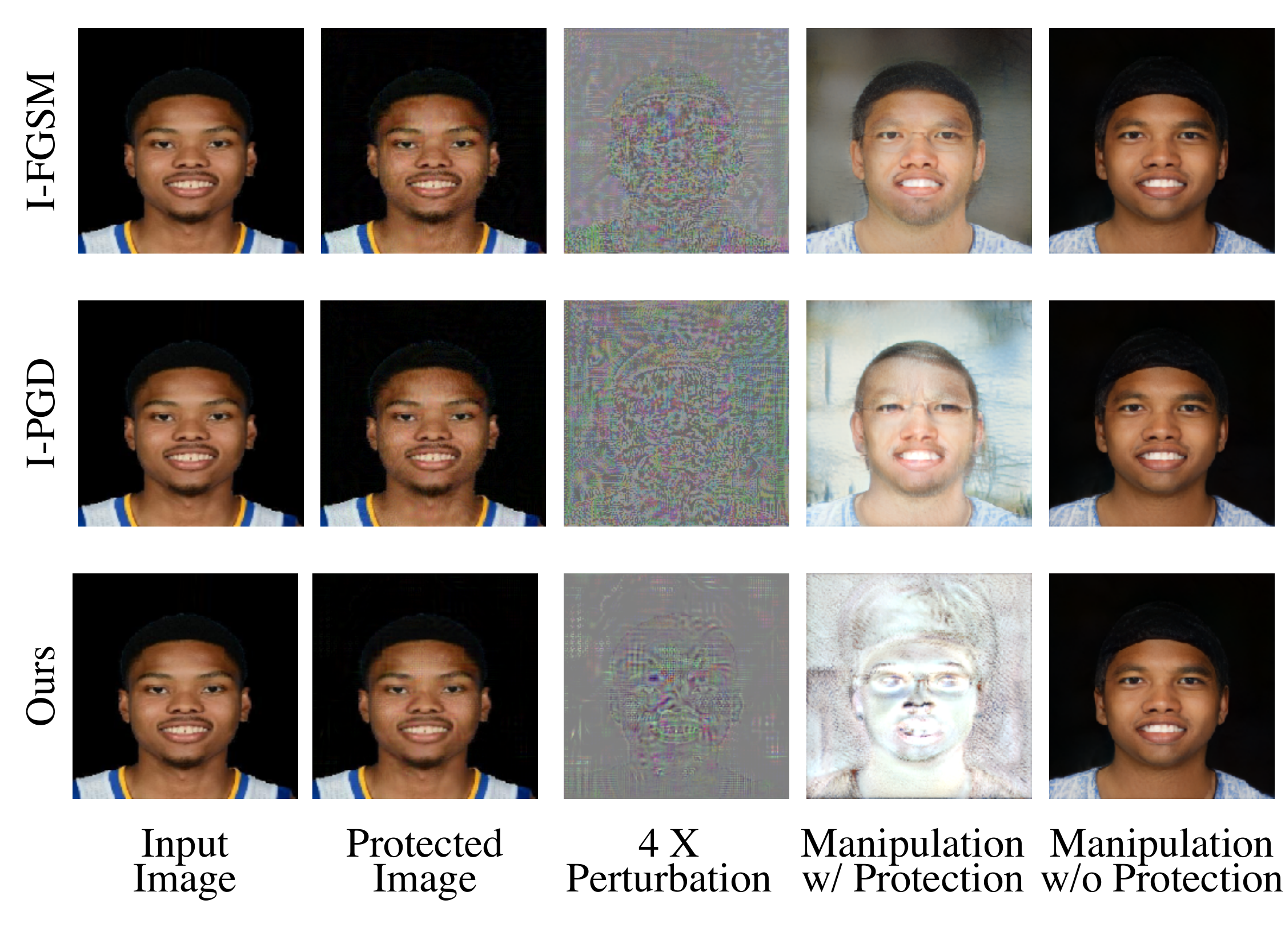}
\end{center}
\caption{Outlier case for the self-reconstruction task~\cite{richardson2021encoding} with white image as manipulation target. In some cases, I-FGSM~\cite{Kurakin2017AdversarialML} and I-PGD~\cite{Madry2018TowardsDL} produce outliers with high mean squared error;~cf. the high variance compared to our method in Fig.~\ref{fig:self_recon_violin_plot}.}
\label{fig:high_var_ex}
\end{figure}

We analyze the output distribution of the manipulation model for different methods in Fig.~\ref{fig:self_recon_violin_plot}. We notice that per-image optimization techniques, like I-PGD~\cite{Madry2018TowardsDL} show a number of outliers with high mean squared error in the generated output images, whereas our proposed method shows quite less variance. One such outlier is visualized in Fig.~\ref{fig:high_var_ex}.

\subsection{Ablations with different Norms: $\boldsymbol{L_{1}}$, $\boldsymbol{L_{2}}$, $\boldsymbol{L_{\infty}}$}\label{sec:diff_norms}
\begin{figure}[h!]
\begin{center}
\includegraphics[width=1.0\linewidth]{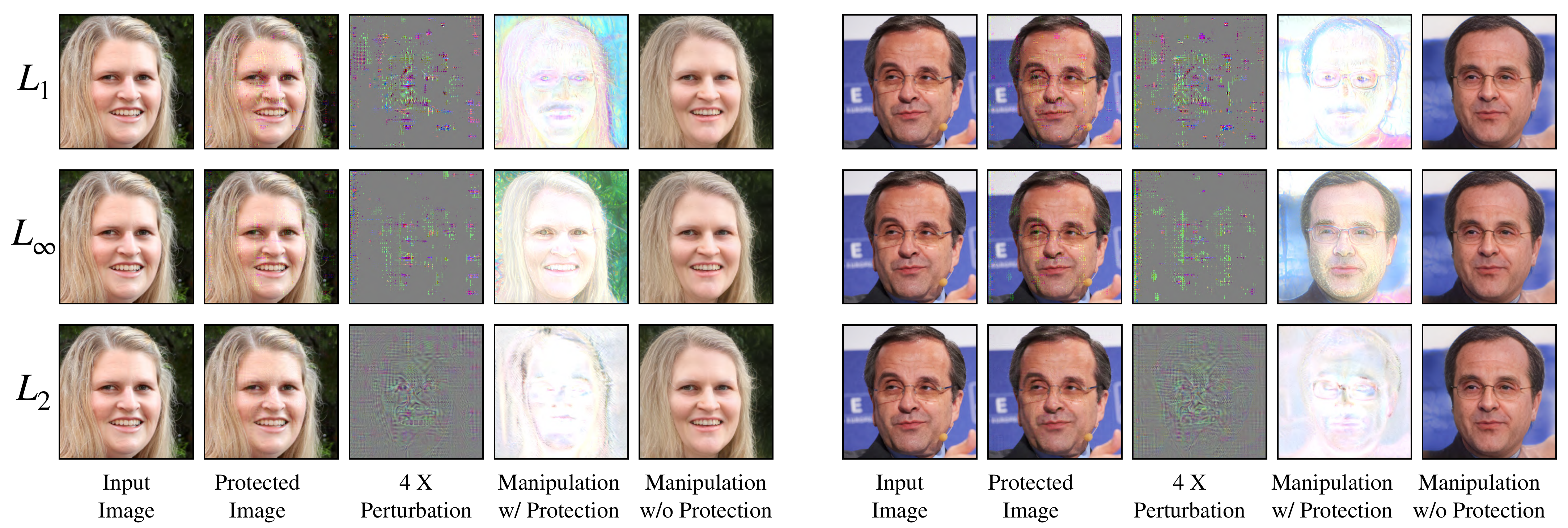}
\end{center}
\caption{Results for self-reconstruction task with white image as manipulation target. Comparison for three different norms used for the loss function: $L_1, L_2 \text{ and } L_{\infty}$. $L_2$ norm evenly distributes the perturbation over the image making it less perceptually visible. Also, the output images are more similar to white manipulation target.}
\label{fig:visual_diff_norms}
\end{figure}

\begin{figure}[h!]
\begin{center}
\includegraphics[width=0.5\linewidth]{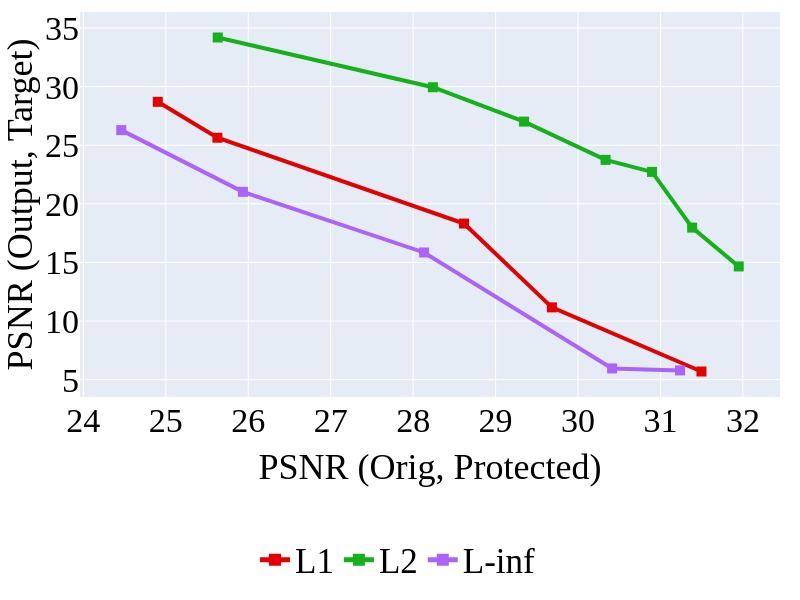}
\end{center}
\caption{Results for self-reconstruction task with white image as manipulation target. Performance comparison on PSNR graph for different norms. Orig and Protected refer to the original and protected image. Output refers to the output of the manipulation model and Target indicates predefined manipulation target. Note that $L_2$ norm outperforms $L_1$ and $L_{\infty}$ based formulations.}
\label{fig:self_recon_norm_ablations}
\end{figure}

Our loss function/minimization objective is formulated in terms of the $L_2$ norm. In this section, we compare results of our method with $L_1$ and $L_{\infty}$ norm respectively. Visual results are shown in Fig.~\ref{fig:visual_diff_norms} and performance comparison in Fig.~\ref{fig:self_recon_norm_ablations}. We notice that $L_2$ norm significantly outperforms the other norms evenly distributing perturbation throughout the image making the changes in the image much less perceptually visible. In addition, it is more effective in erasing the image traces from the output of manipulation model.

\subsection{Multi-level Compression}\label{sec:multi_level_comp}
\begin{figure*}[h!]
\begin{center}
\includegraphics[width=0.85\linewidth]{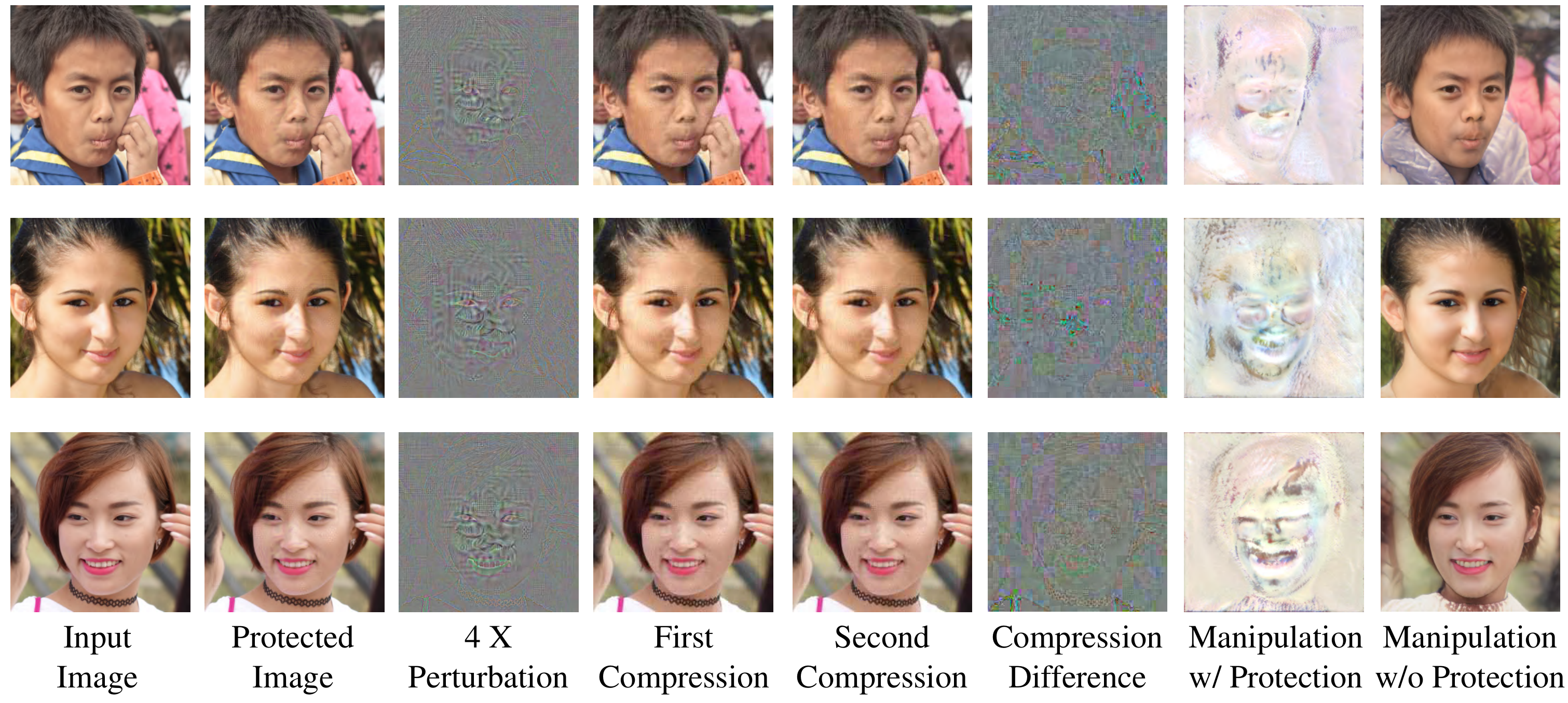}
\end{center}
\caption{Results for the self-reconstruction task with white image as manipulation target and for multi-level compression. First Compression denotes the first applied high compression C-30, Second Compression denotes the second compression (C-80) applied to the first compressed image. We apply different compression to evaluate robustness. Compression Difference denotes the difference between first and second compressed images amplified by 20X for visibility. Our method can efficiently handle multiple compression levels of different qualities.}
\label{fig:multi_level_comp}
\end{figure*}

In a practical use case, an image might be compressed multiple times when shared on social media platforms. Therefore, the protection applied to images should be robust to consecutively applied compression. For simplicity, we show results for bi-level compression, i.e. the compression is applied twice to the image. We first apply a high compression (C-30) and thereafter a low compression (C-80) for evaluation. The second compression is applied to the first compressed image. Results are shown in Fig.~\ref{fig:multi_level_comp}. We show that our method is robust to multi-level compression as well.

\subsection{Manipulation for higher perturbation levels}\label{sec:high_comp}
\begin{figure}[h!]
\begin{center}
\includegraphics[width=1.0\linewidth]{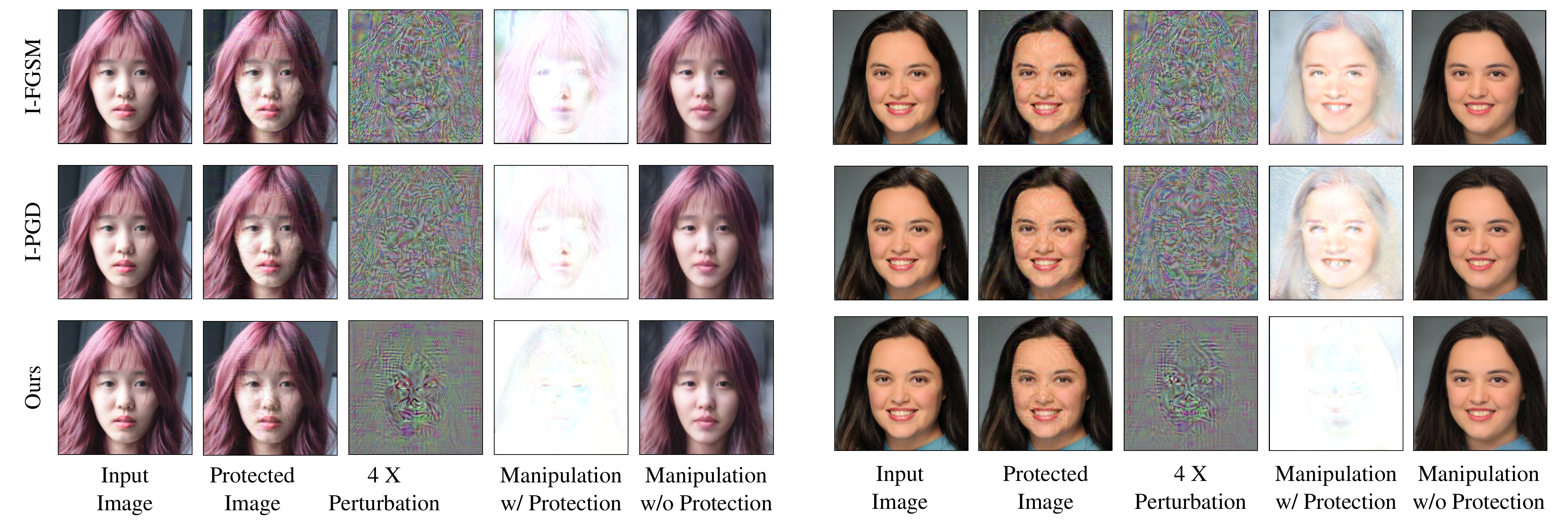}
\end{center}
\caption{Visual results for higher perturbation levels for the self-reconstruction task~\cite{richardson2021encoding} with white image as manipulation target. Even for a higher perturbation level, our proposed approach outperforms alternate methods.}
\label{fig:high_comp}
\end{figure}

We show visual comparisons for lower perturbation levels since these are most useful for practical purposes. To this end, in the graphs shown in the main paper, we have analyzed our method at several different perturbation levels. In this section, we visualize our results for higher perturbation levels, which show more disturbance in the original images, but illustrated more significant disruptions for manipulation model predictions. Fig.~\ref{fig:high_comp} shows the visual results for the self-reconstruction task.

\subsection{Ablations with different JPEG approximations}\label{sec:diff_jpeg_approxs}

There are three different approaches to approximate the $\round$ operation used in the quantization step of the original JPEG compression technique. These are formalized below.

\begin{figure}[h!]
\begin{center}
\includegraphics[width=0.5\linewidth]{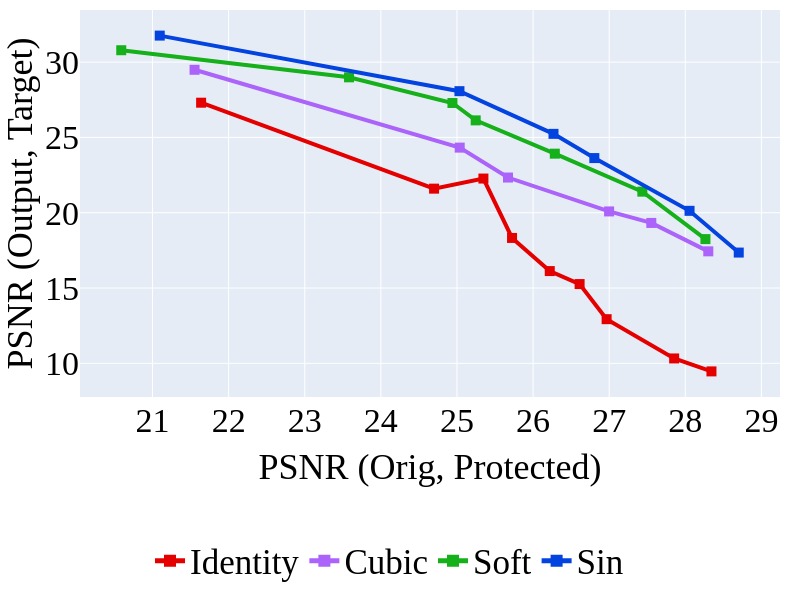}
\end{center}
\caption{PSNR performance for training our method with different approximations to round operation for JPEG compression for the self-reconstruction task~\cite{richardson2021encoding} with white image as manipulation target. Sin approximation outperforms other approximations with comparable results for Soft approximation.}
\label{fig:jpeg_ablations}
\end{figure}

\begin{enumerate}
    \item Cubic Approximation ~\cite{Shin2017JPEGresistantAI}
    $$
    x:= \lfloor x \rceil + \big(x - \lfloor x \rceil \big)^3.
    $$
    \item Soft Approximation~\cite{Korus_2019_CVPR}
    $$
  \tilde{x} =  x  - \dfrac{\sin(2\pi x)}{2\pi},
$$

$$
  x:= \big[ \round(x) - \tilde{x} \big]_{\text{detach}} + \tilde{x},
$$
where "$\text{detach}$" indicates that no gradients will be propagated during backpropagation.
\item Sin Approximation~\cite{Korus_2019_CVPR}
$$
    {x} :=  x  - \dfrac{\sin(2\pi x)}{2\pi}
$$
\end{enumerate}

We also compare against the identity operation $x:= x$ as the baseline. The performance comparison for these methods is visualized in Fig.~\ref{fig:jpeg_ablations}. We notice that Sin approximation is better in performance compared to other approximations, therefore we use it for experiments in the main paper.

\clearpage

%
%
\bibliographystyle{splncs04}
\bibliography{egbib}
\end{document}